\documentclass{article}

\usepackage[preprint]{neurips_2025}

\usepackage[utf8]{inputenc} 
\usepackage[T1]{fontenc}    
\usepackage{url}            
\usepackage{booktabs}       
\usepackage{amsfonts}       
\usepackage{nicefrac}       
\usepackage{microtype}      
\usepackage{graphicx}
\usepackage{amsmath}
\usepackage{wrapfig}
\usepackage{multirow}
\usepackage[table]{xcolor}
\usepackage{makecell}
\usepackage{algorithm}
\usepackage{algpseudocode}
\definecolor{lightorange}{RGB}{255, 236, 204}

\definecolor{scholarblue}{rgb}{0.21,0.49,0.74}
\usepackage[pagebackref,breaklinks,colorlinks]{hyperref}
\hypersetup{
	colorlinks=true,
	linkcolor=scholarblue,
	filecolor=blue,      
	urlcolor=scholarblue,
	citecolor=scholarblue,
}

\usepackage{listings}
\usepackage{color}

\newcommand{\ie}{i.e.,~}
\newcommand{\eg}{\emph{e.g.},~}

\newcommand{\normal}[1]{\ensuremath{\mathcal{N}\left(#1\right)}}

\newcommand{\change}[1]{\textcolor{blue}{#1}}

\newcommand{\bv}{{\mathbf{v}}}

\newcommand{\bx}{{\mathbf{x}}}

\newcommand{\bz}{{\mathbf{z}}}

\newcommand{\bI}{{\mathbf{I}}}

\newcommand{\btheta}{\mathbf{\theta}}

\newcommand{\bepsilon}{{\boldsymbol{\epsilon}}}

\newcommand{\figref}[1]{Figure~\ref{#1}}

\newcommand{\secref}[1]{Section~\ref{#1}}

\newcommand{\tabref}[1]{Table~\ref{#1}}

\usepackage{mdframed}
\definecolor{mdproofbg}{rgb}{0.95,0.95,0.95}
{\begin{mdframed}[backgroundcolor=mdproofbg,linewidth=0]\begin{proof}}%
{\end{proof}\end{mdframed}}

\definecolor{mdworkingbg}{rgb}{1.0,0.95,0.95}
{\begin{mdframed}[backgroundcolor=mdworkingbg,linewidth=0]\begin{minipage}{\columnwidth}}%
{\end{minipage}\end{mdframed}}

\usepackage{xpatch}
\makeatletter
\AtBeginDocument{\xpatchcmd{\@thm}{\thm@headpunct{.}}{\thm@headpunct{:}}{}{}} 
\AtBeginDocument{\xpatchcmd{\@thm}{\fontseries\mddefault\upshape}{}{}{}} 
\makeatother

\definecolor{bgCode}{rgb}{0.98, 0.98, 0.98}
\definecolor{codegray}{rgb}{0.5,0.5,0.5}
\lstnewenvironment{pycode}%
{  \noindent
	\minipage{\linewidth} 
	\bigskip 
	\lstset{language=Python,numbers=left,numbersep=5pt}
	\raggedright
	\scriptsize{\textsf{Python Code}\vspace{-1.75ex}}
	\raggedleft}%
{\smallskip \endminipage}

\title{DiSA: Diffusion Step Annealing in Autoregressive Image Generation}

\author{Qinyu Zhao\textsuperscript{1}, Jaskirat Singh\textsuperscript{1}, Ming Xu\textsuperscript{1}, Akshay Asthana\textsuperscript{2}, Stephen Gould\textsuperscript{1}, Liang Zheng\textsuperscript{1} \\
\textsuperscript{1} Australian National University\\
\textsuperscript{2} Seeing Machines Ltd\\
\texttt{\{qinyu.zhao,jaskirat.singh,mingda.xu,stephen.gould,liang.zheng\}@anu.edu.au} \\
\texttt{\{akshay.asthana\}@seeingmachines.com}\\
}

\begin{document}

\maketitle

\begin{abstract}
An increasing number of autoregressive models, such as MAR, FlowAR, xAR, and Harmon adopt diffusion sampling to improve the quality of image generation. However, this strategy leads to low inference efficiency, because it usually takes 50 to 100 steps for diffusion to sample a token. This paper explores how to effectively address this issue.
Our key motivation is that as more tokens are generated during the autoregressive process, subsequent tokens follow more constrained distributions and are easier to sample. To intuitively explain, if a model has generated part of a dog, the remaining tokens must complete the dog and thus are more constrained. Empirical evidence supports our motivation: at later generation stages, the next tokens can be well predicted by a multilayer perceptron, exhibit low variance, and follow closer-to-straight-line denoising paths from noise to tokens. 
Based on our finding, we introduce diffusion step annealing (DiSA), a training-free method which gradually uses fewer diffusion steps as more tokens are generated, \eg using 50 steps at the beginning and gradually decreasing to 5 steps at later stages. Because DiSA is derived from our finding specific to diffusion in autoregressive models, it is complementary to existing acceleration methods designed for diffusion alone. 
DiSA can be implemented in only a few lines of code on existing models, and albeit simple, achieves $5-10\times$ faster inference for MAR and Harmon and $1.4-2.5\times$ for FlowAR and xAR, while maintaining the generation quality.
\end{abstract}

\section{Introduction}\label{sec:introduction}
\begin{figure}[t]
  \centering
  \includegraphics[width=0.98\textwidth,trim=0 130 135 0,clip]{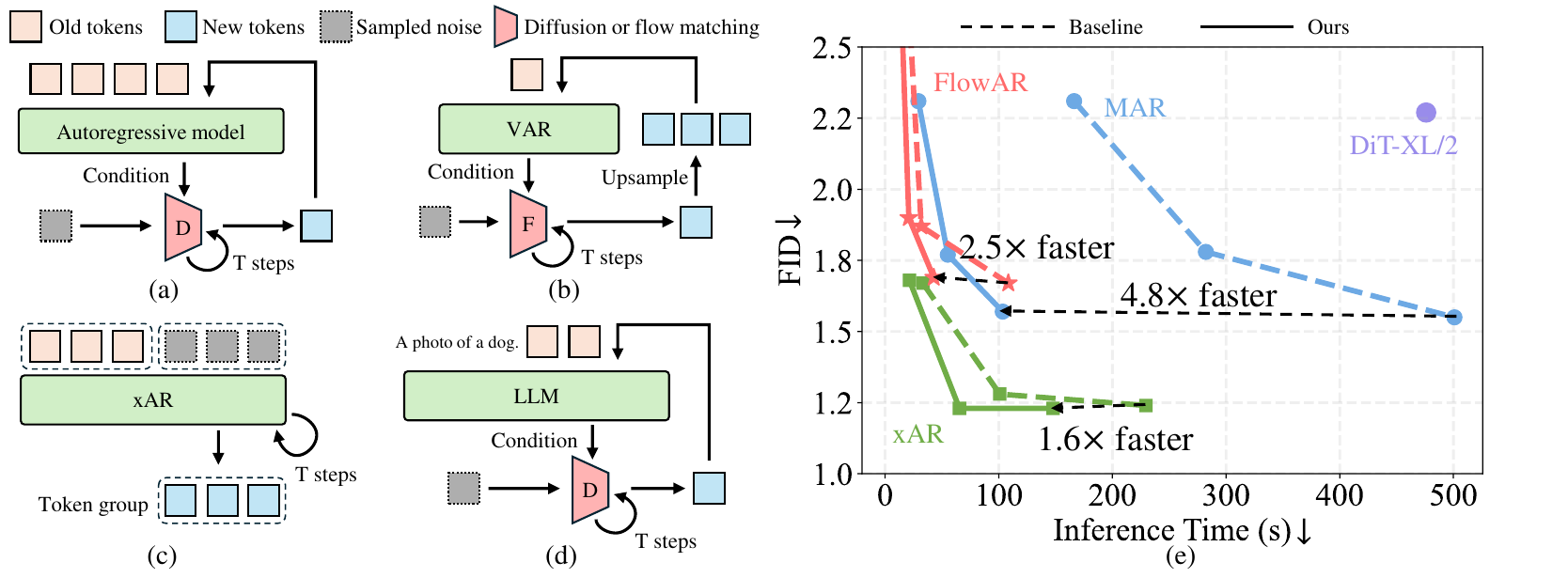}
  \caption{\textbf{Overview}. Architecture of four ``autoregressive + diffusion'' models included in this study: (a) MAR~\cite{mar}; (b) FlowAR~\cite{ren2024flowar}; (c) xAR~\cite{xar}; (d) Harmon~\cite{harmon}. (e) This paper improves the efficiency of these models by reducing diffusion steps without compromising generation quality.}
  \label{fig:framework}
  \vskip -0.1in
\end{figure}

A growing number of autoregressive models introduce diffusion sampling to generate continuous tokens, such as MAR~\cite{mar}, FlowAR~\cite{ren2024flowar}, xAR~\cite{xar}, and Harmon~\cite{harmon}, which significantly improves generation quality. At inference time, these models take autoregressively generated tokens as input and adopt a diffusion process to sample the next tokens. An illustration is shown in \figref{fig:framework}(a-d).

Although the diffusion process yields higher image quality for autoregressve models, it suffers from low inference efficiency because tens of denosing steps are needed to generate each token. For example, MAR~\cite{mar} denoises 100 times while xAR~\cite{xar} does 50 times. Our preliminary experiments show that the many-step diffusion process accounts for about 50\% inference latency in MAR and 90\% in xAR. Naively reducing the number of diffusion steps would accelerate these models but will significantly degrade generation quality. For example, with 10 diffusion steps, the Fréchet Inception Distance~(FID) of xAR-L on ImageNet 256$\times$256 increases shapely from 1.28 to 8.6, and MAR-L even fails to generate meaningful images. This paper thus aims to address the efficiency issue.

We are motivated by the finding that as more tokens are generated, token distributions become more constrained, and tokens become easier to sample. In other words, early generation stages are reliant on stronger distribution modeling and token sampling, while late stages are less so. 

\begin{figure*}
\centering
\renewcommand\arraystretch{0.6}
\renewcommand\tabcolsep{0pt}
\begin{tabular}{lccccccccc}
\multirow{5}[9]{*}{\rotatebox{90}{MAR-L}} &
\hspace{2pt}\includegraphics[width=.10\textwidth]{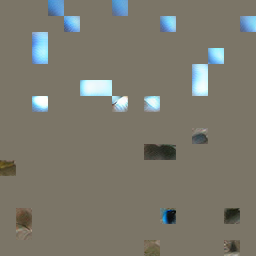}
&\includegraphics[width=.10\textwidth]{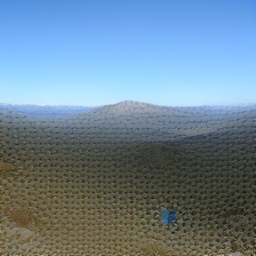}
&\hspace{2pt}\includegraphics[width=.10\textwidth]{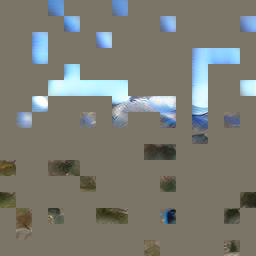}
&\includegraphics[width=.10\textwidth]{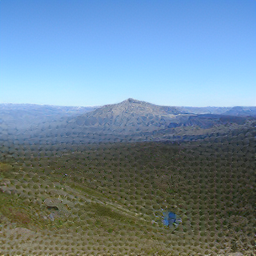}
&\hspace{2pt}\includegraphics[width=.10\textwidth]{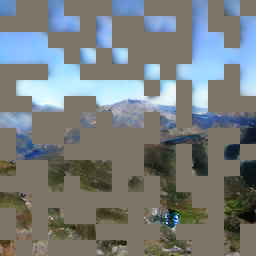}
&\includegraphics[width=.10\textwidth]{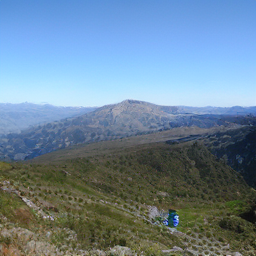}
&\hspace{2pt}\includegraphics[width=.10\textwidth]{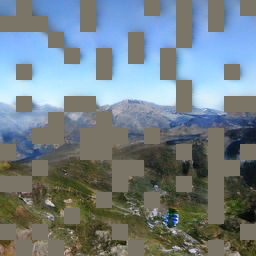}
&\includegraphics[width=.10\textwidth]{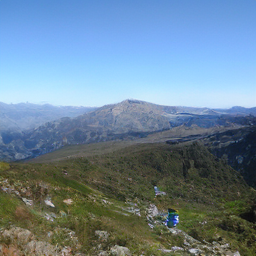} & \hspace{2pt}\includegraphics[width=.10\textwidth]{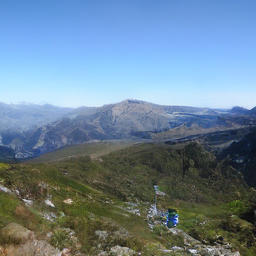}
\\
& \hspace{2pt}\includegraphics[width=.10\textwidth]{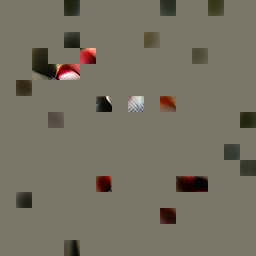}
&\includegraphics[width=.10\textwidth]{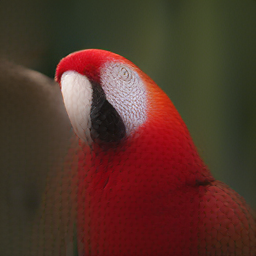}
&\hspace{2pt}\includegraphics[width=.10\textwidth]{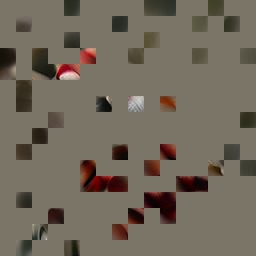}
&\includegraphics[width=.10\textwidth]{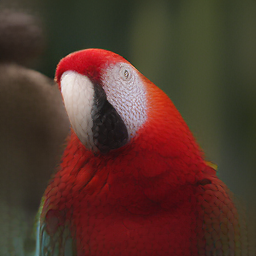}
&\hspace{2pt}\includegraphics[width=.10\textwidth]{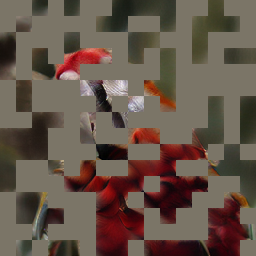}
&\includegraphics[width=.10\textwidth]{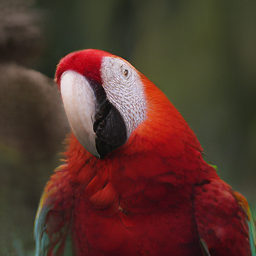}
&\hspace{2pt}\includegraphics[width=.10\textwidth]{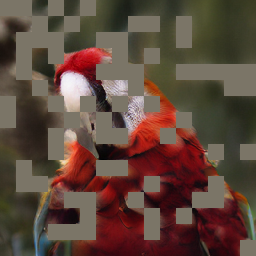}
&\includegraphics[width=.10\textwidth]{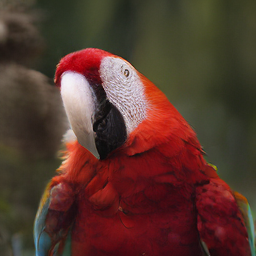} & 
\hspace{2pt}\includegraphics[width=.10\textwidth]{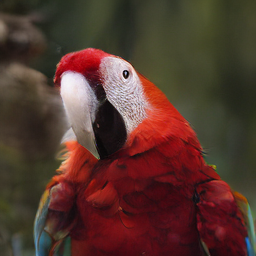}
\\
 
& \hspace{2pt}\includegraphics[width=.10\textwidth]{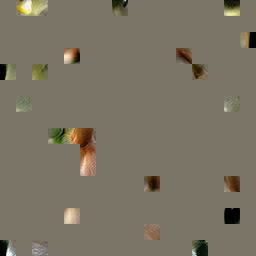}
&\includegraphics[width=.10\textwidth]{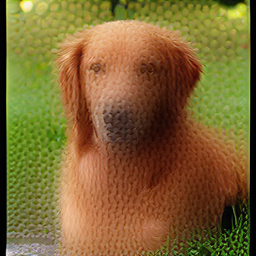}
&\hspace{2pt}\includegraphics[width=.10\textwidth]{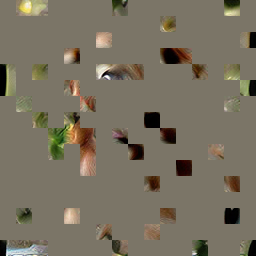}
&\includegraphics[width=.10\textwidth]{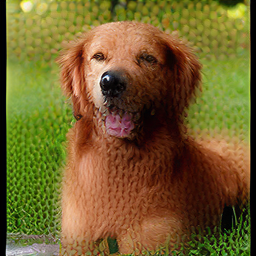}
&\hspace{2pt}\includegraphics[width=.10\textwidth]{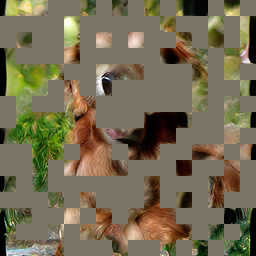}
&\includegraphics[width=.10\textwidth]{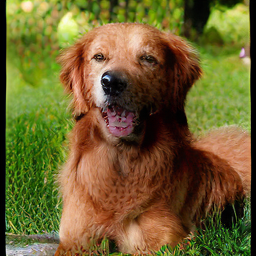}
&\hspace{2pt}\includegraphics[width=.10\textwidth]{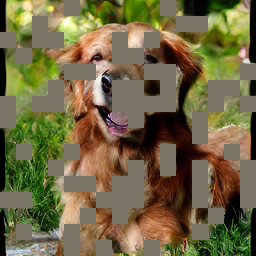}
&\includegraphics[width=.10\textwidth]{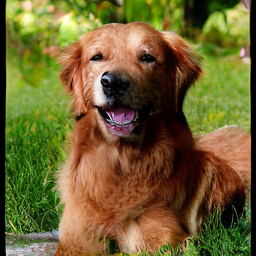}
&\hspace{2pt}\includegraphics[width=.10\textwidth]{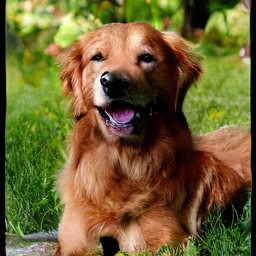}
\\
& \multicolumn{2}{c}{AR Step 18} & \multicolumn{2}{c}{AR Step 26} & \multicolumn{2}{c}{AR Step 42} &\multicolumn{2}{c}{AR Step 53} & Generated\\

\multirow{5}[9]{*}{\rotatebox{90}{FlowAR-L}} &
\hspace{2pt}\includegraphics[width=.10\textwidth]{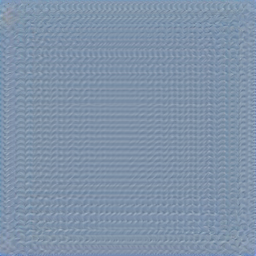}
&\includegraphics[width=.10\textwidth]{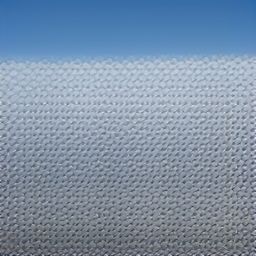}
&\hspace{2pt}\includegraphics[width=.10\textwidth]{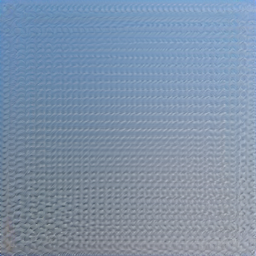}
&\includegraphics[width=.10\textwidth]{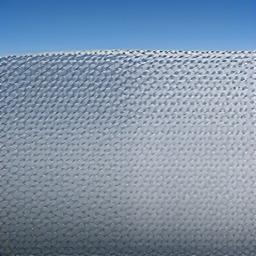}
&\hspace{2pt}\includegraphics[width=.10\textwidth]{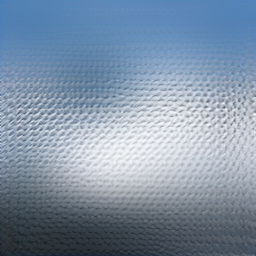}
&\includegraphics[width=.10\textwidth]{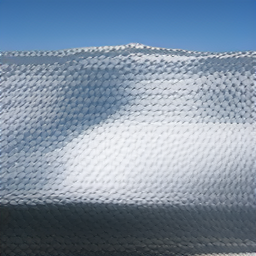}
&\hspace{2pt}\includegraphics[width=.10\textwidth]{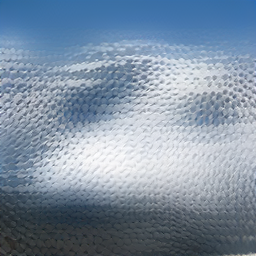}
&\includegraphics[width=.10\textwidth]{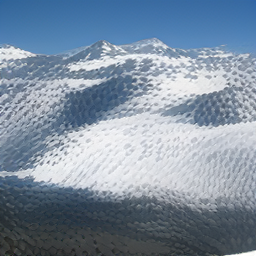} & \hspace{2pt}\includegraphics[width=.10\textwidth]{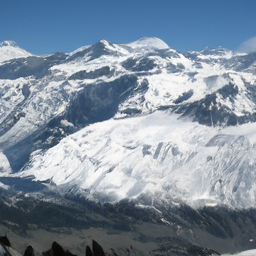}
\\
& \hspace{2pt}\includegraphics[width=.10\textwidth]{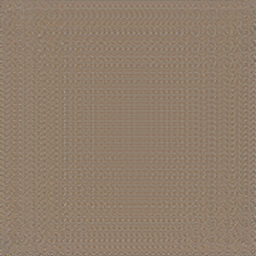}
&\includegraphics[width=.10\textwidth]{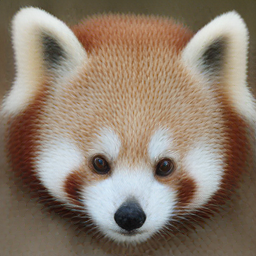}
&\hspace{2pt}\includegraphics[width=.10\textwidth]{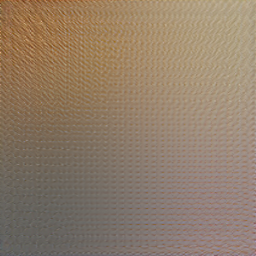}
&\includegraphics[width=.10\textwidth]{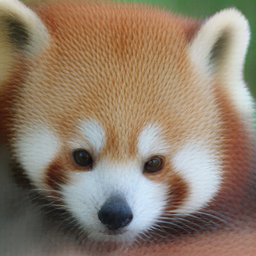}
&\hspace{2pt}\includegraphics[width=.10\textwidth]{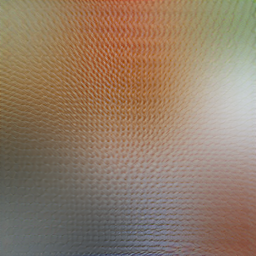}
&\includegraphics[width=.10\textwidth]{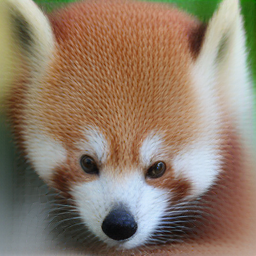}
&\hspace{2pt}\includegraphics[width=.10\textwidth]{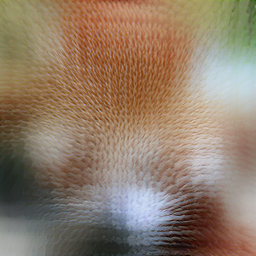}
&\includegraphics[width=.10\textwidth]{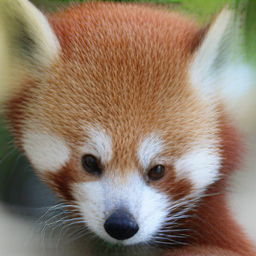} & 
\hspace{2pt}\includegraphics[width=.10\textwidth]{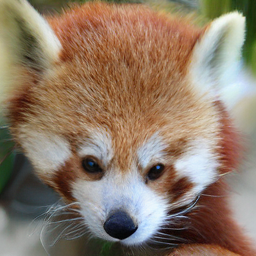}
\\
& \hspace{2pt}\includegraphics[width=.10\textwidth]{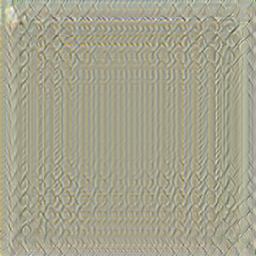}
&\includegraphics[width=.10\textwidth]{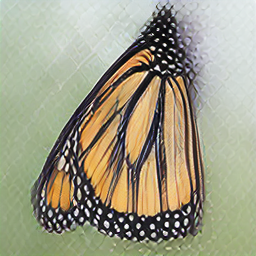}
&\hspace{2pt}\includegraphics[width=.10\textwidth]{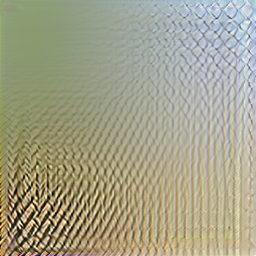}
&\includegraphics[width=.10\textwidth]{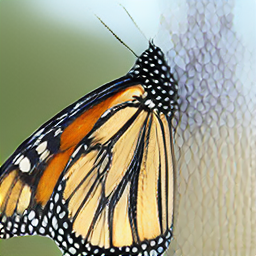}
&\hspace{2pt}\includegraphics[width=.10\textwidth]{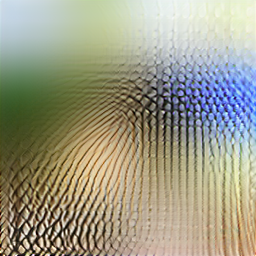}
&\includegraphics[width=.10\textwidth]{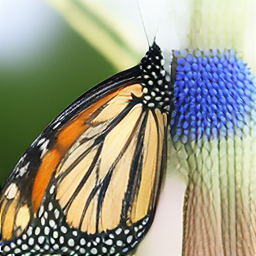}
&\hspace{2pt}\includegraphics[width=.10\textwidth]{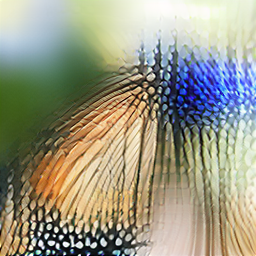}
&\includegraphics[width=.10\textwidth]{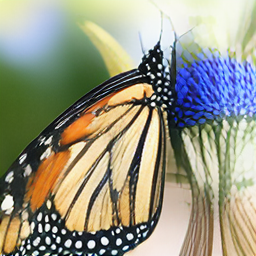}
&\hspace{2pt}\includegraphics[width=.10\textwidth]{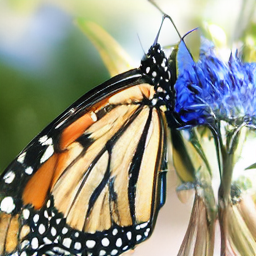}
\\
& \multicolumn{2}{c}{AR Step 1} & \multicolumn{2}{c}{AR Step 2} & \multicolumn{2}{c}{AR Step 3} &\multicolumn{2}{c}{AR Step 4} & Generated \\

\multirow{5}[9]{*}{\rotatebox{90}{xAR-L}} &
\hspace{2pt}\includegraphics[width=.10\textwidth]{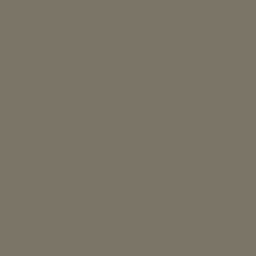}
&\includegraphics[width=.10\textwidth]{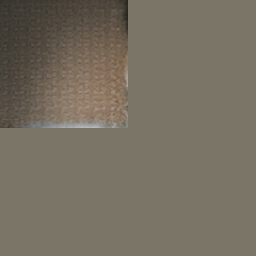}
&\hspace{2pt}\includegraphics[width=.10\textwidth]{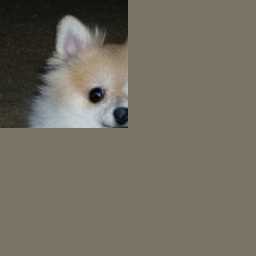}
&\includegraphics[width=.10\textwidth]{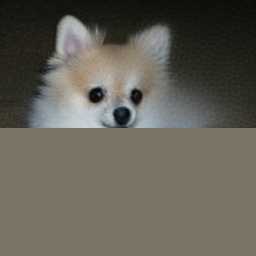}
&\hspace{2pt}\includegraphics[width=.10\textwidth]{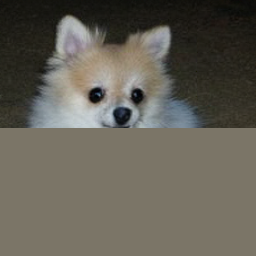}
&\includegraphics[width=.10\textwidth]{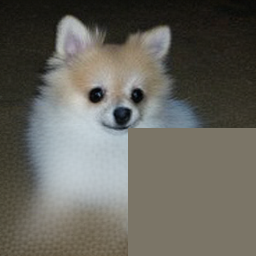}
&\hspace{2pt}\includegraphics[width=.10\textwidth]{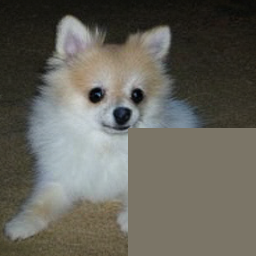}
&\includegraphics[width=.10\textwidth]{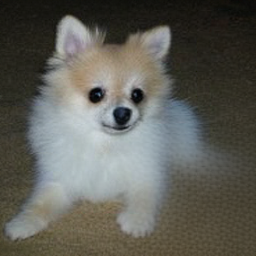} & \hspace{2pt}\includegraphics[width=.10\textwidth]{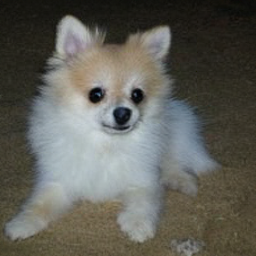}
\\
& \hspace{2pt}\includegraphics[width=.10\textwidth]{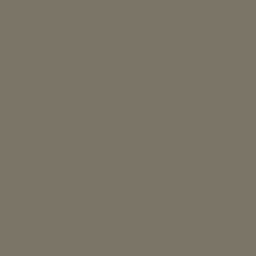}
&\includegraphics[width=.10\textwidth]{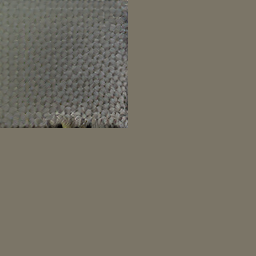}
&\hspace{2pt}\includegraphics[width=.10\textwidth]{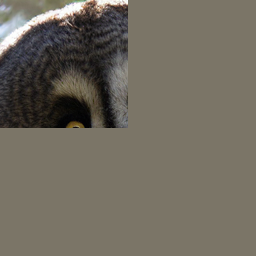}
&\includegraphics[width=.10\textwidth]{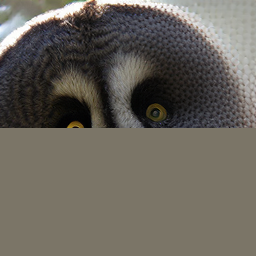}
&\hspace{2pt}\includegraphics[width=.10\textwidth]{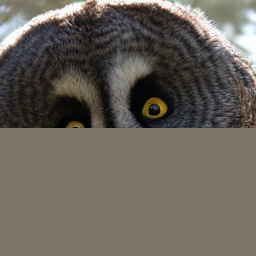}
&\includegraphics[width=.10\textwidth]{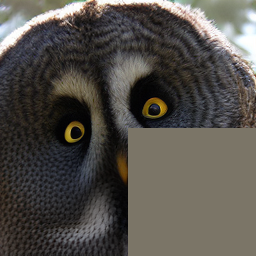}
&\hspace{2pt}\includegraphics[width=.10\textwidth]{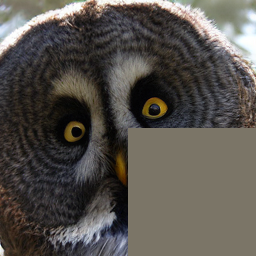}
&\includegraphics[width=.10\textwidth]{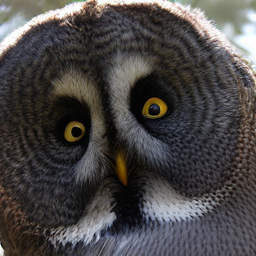} & 
\hspace{2pt}\includegraphics[width=.10\textwidth]{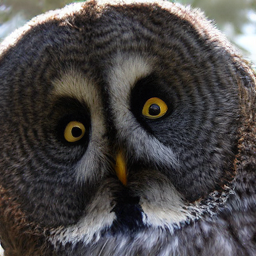}
\\
& \hspace{2pt}\includegraphics[width=.10\textwidth]{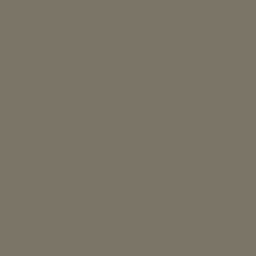}
&\includegraphics[width=.10\textwidth]{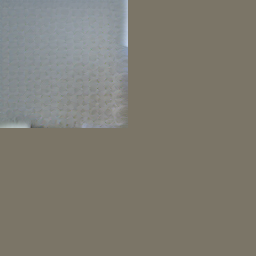}
&\hspace{2pt}\includegraphics[width=.10\textwidth]{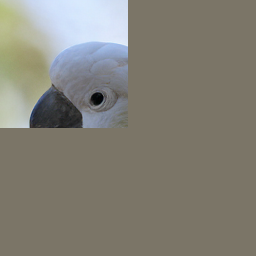}
&\includegraphics[width=.10\textwidth]{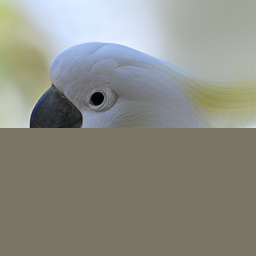}
&\hspace{2pt}\includegraphics[width=.10\textwidth]{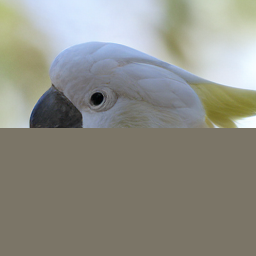}
&\includegraphics[width=.10\textwidth]{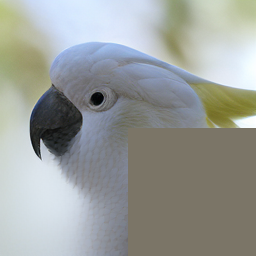}
&\hspace{2pt}\includegraphics[width=.10\textwidth]{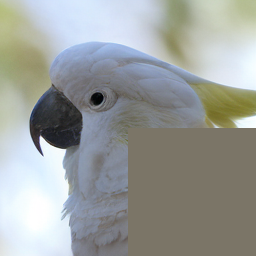}
&\includegraphics[width=.10\textwidth]{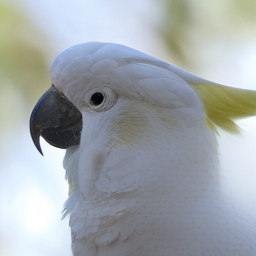}
&\hspace{2pt}\includegraphics[width=.10\textwidth]{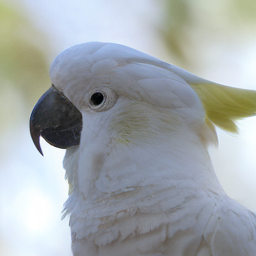}
\\
& \multicolumn{2}{c}{AR Step 0} & \multicolumn{2}{c}{AR Step 1} & \multicolumn{2}{c}{AR Step 2} &\multicolumn{2}{c}{AR Step 3} & Generated \\

& \multicolumn{9}{l}{\scriptsize\textit{\quad Paper artwork, layered paper, colorful Chinese dragon surrounded by clouds.}}\\
\multirow{4}[23]{*}{\rotatebox{90}{Harmon-1.5B}} &
\hspace{2pt}\includegraphics[width=.10\textwidth]{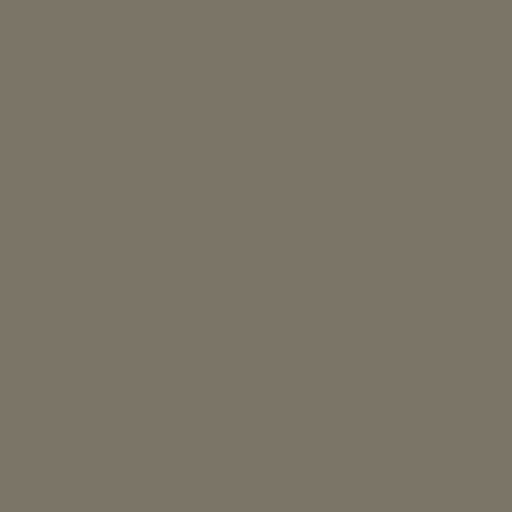}
&\includegraphics[width=.10\textwidth]{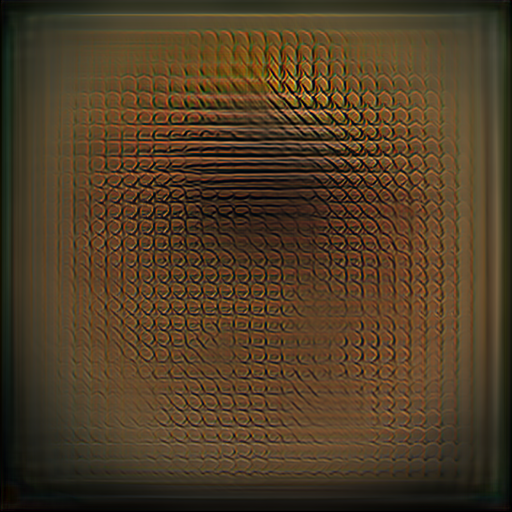}
&\hspace{2pt}\includegraphics[width=.10\textwidth]{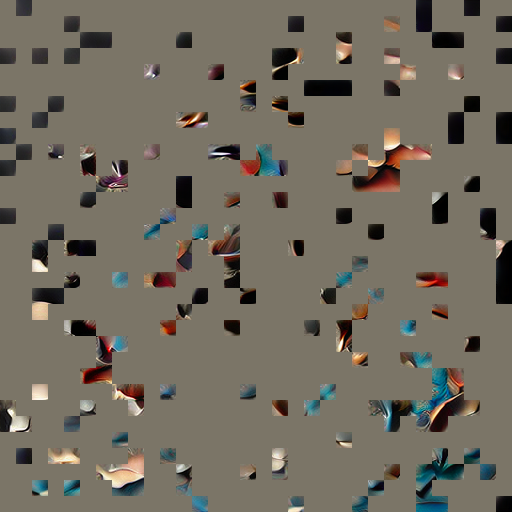}
&\includegraphics[width=.10\textwidth]{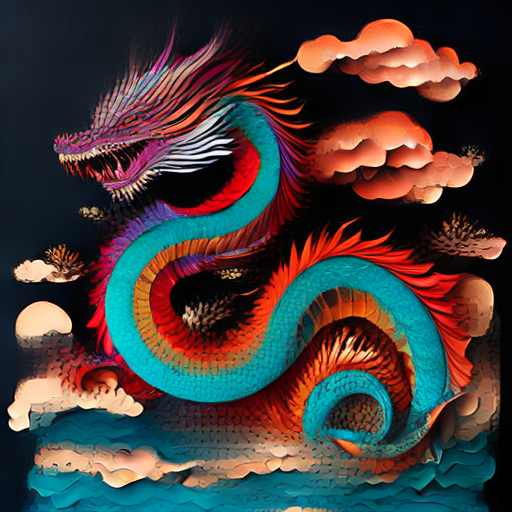}
&\hspace{2pt}\includegraphics[width=.10\textwidth]{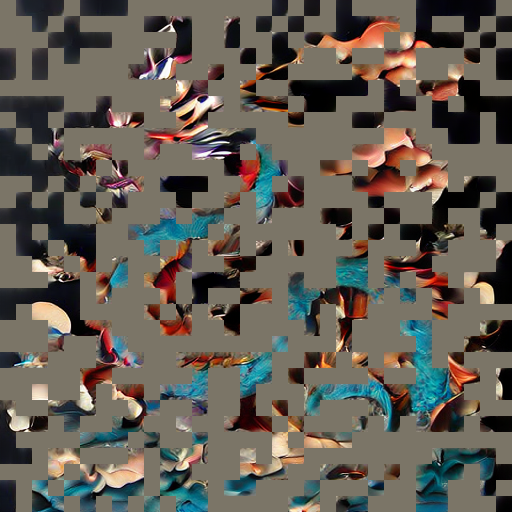}
&\includegraphics[width=.10\textwidth]{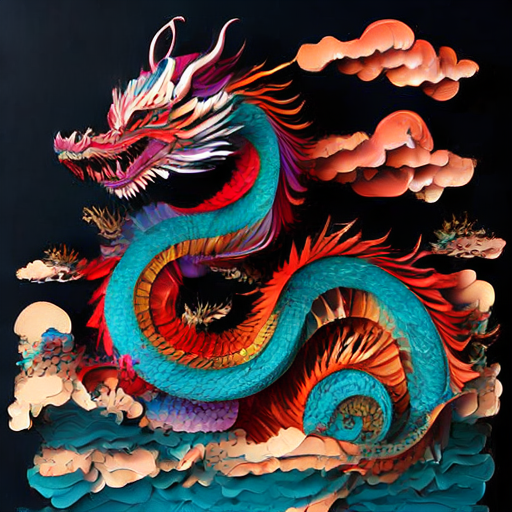}
&\hspace{2pt}\includegraphics[width=.10\textwidth]{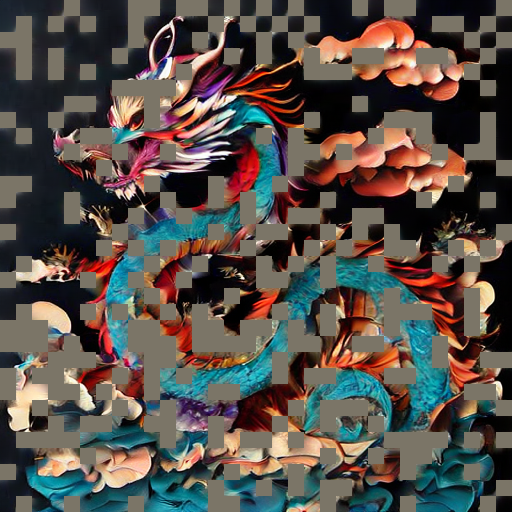}
&\includegraphics[width=.10\textwidth]{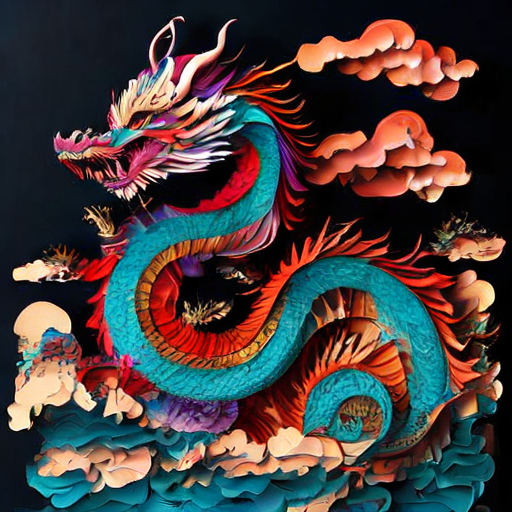}
&\hspace{2pt}\includegraphics[width=.10\textwidth]{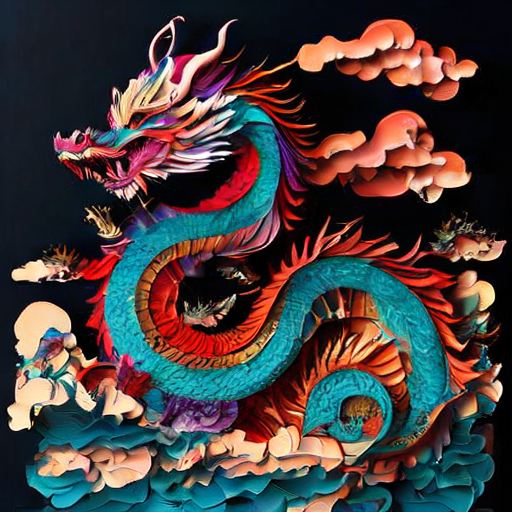}
\\
& \multicolumn{9}{l}{\scriptsize\textit{\quad A photo of a pink stop sign.}}\\
& \hspace{2pt}\includegraphics[width=.10\textwidth]{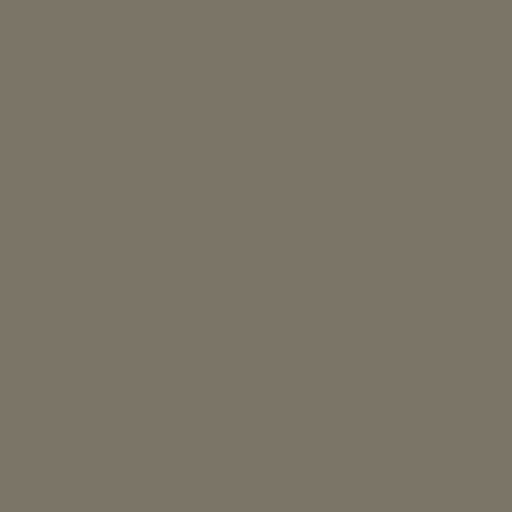}
&\includegraphics[width=.10\textwidth]{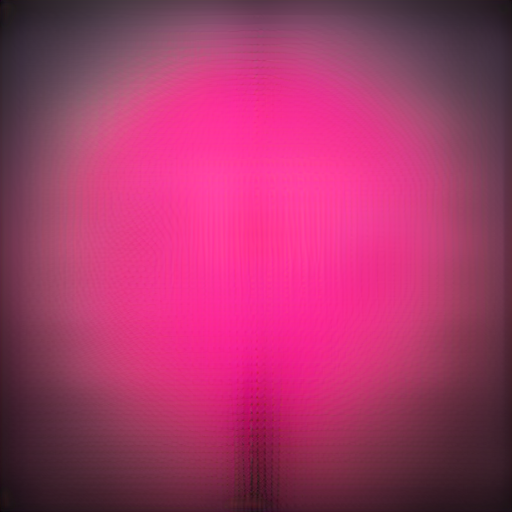}
&\hspace{2pt}\includegraphics[width=.10\textwidth]{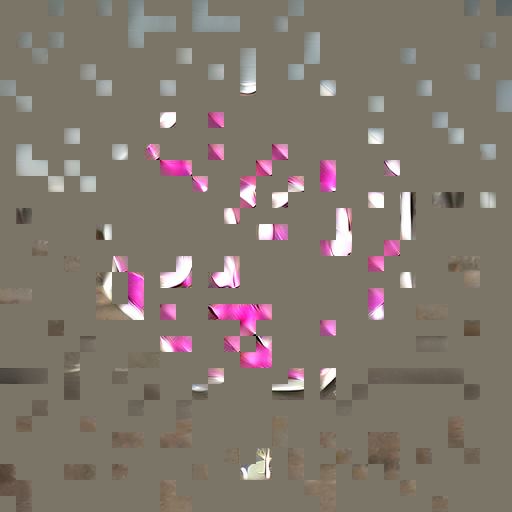}
&\includegraphics[width=.10\textwidth]{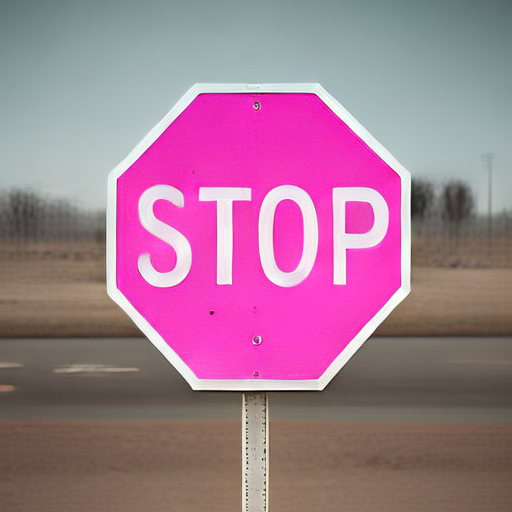}
&\hspace{2pt}\includegraphics[width=.10\textwidth]{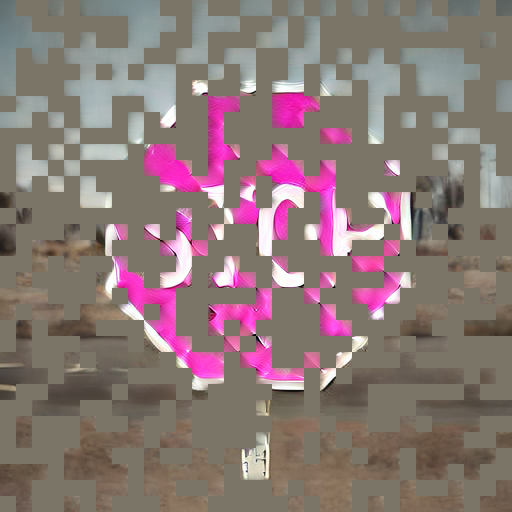}
&\includegraphics[width=.10\textwidth]{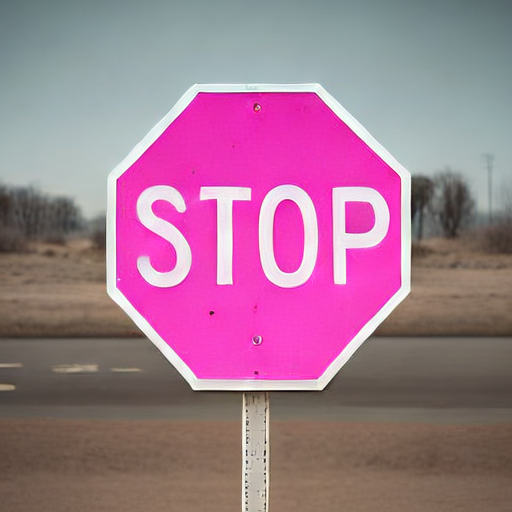}
&\hspace{2pt}\includegraphics[width=.10\textwidth]{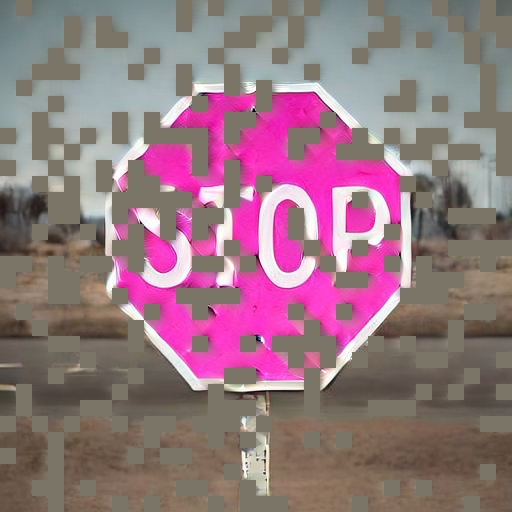}
&\includegraphics[width=.10\textwidth]{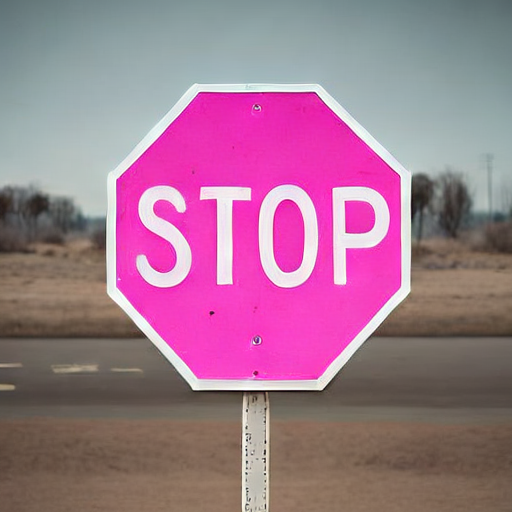}
&\hspace{2pt}\includegraphics[width=.10\textwidth]{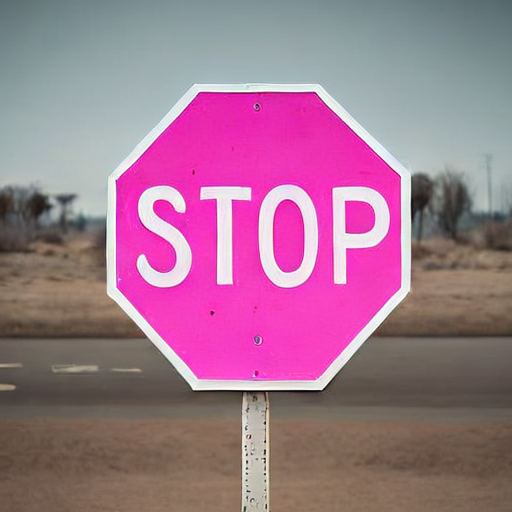}
\\
& \multicolumn{9}{l}{\scriptsize\textit{\quad A realistic landscape shot of the Northern Lights dancing over a snowy mountain range in Iceland.}}\\
& \hspace{2pt}\includegraphics[width=.10\textwidth]{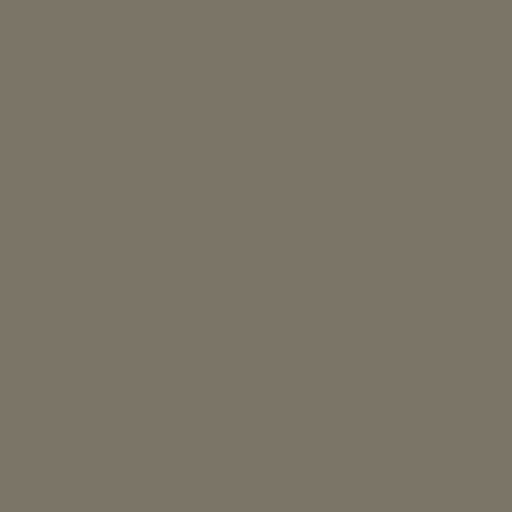}
&\includegraphics[width=.10\textwidth]{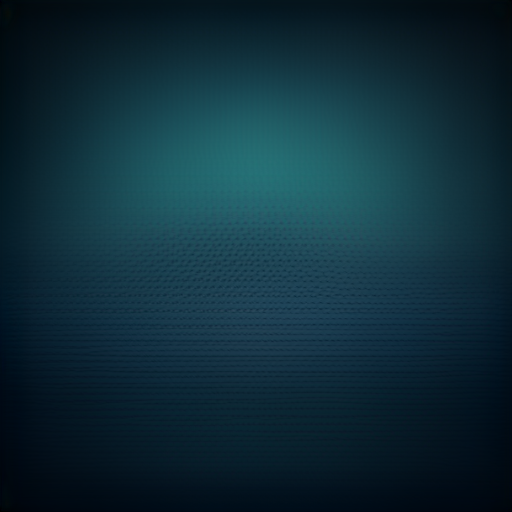}
&\hspace{2pt}\includegraphics[width=.10\textwidth]{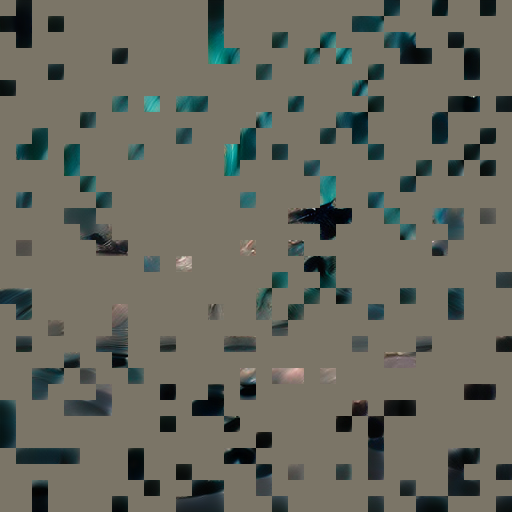}
&\includegraphics[width=.10\textwidth]{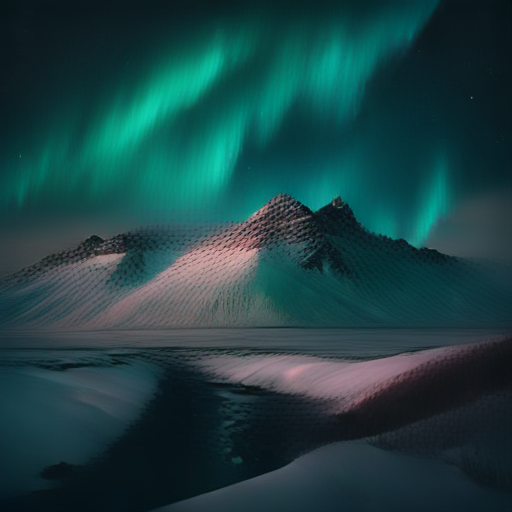}
&\hspace{2pt}\includegraphics[width=.10\textwidth]{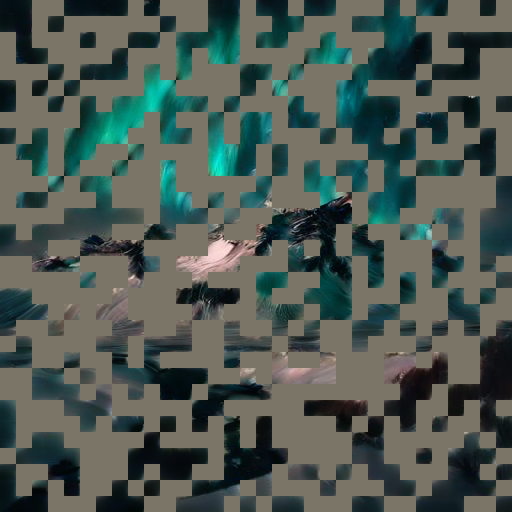}
&\includegraphics[width=.10\textwidth]{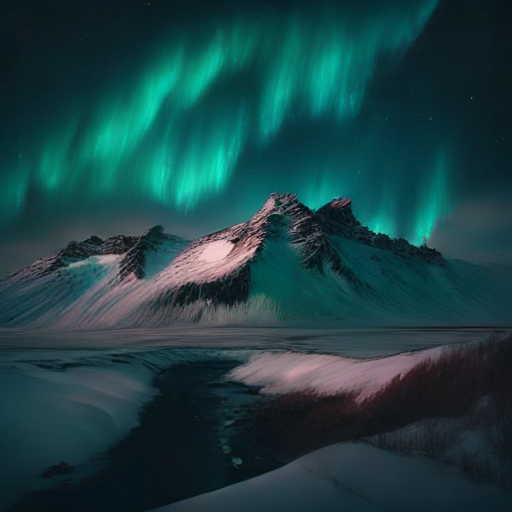}
&\hspace{2pt}\includegraphics[width=.10\textwidth]{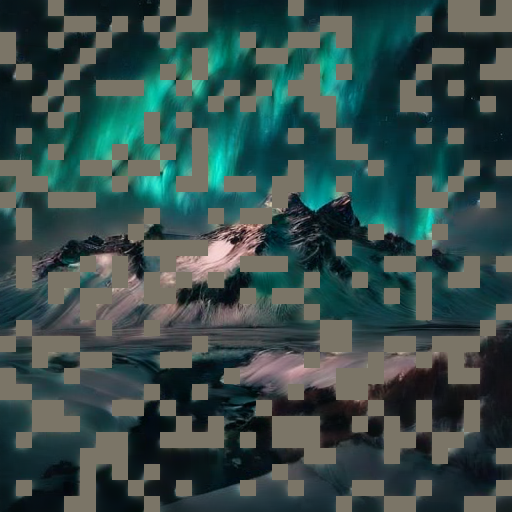}
&\includegraphics[width=.10\textwidth]{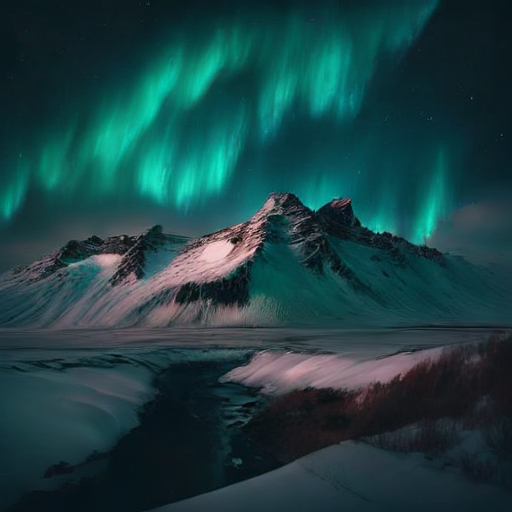}
&\hspace{2pt}\includegraphics[width=.10\textwidth]{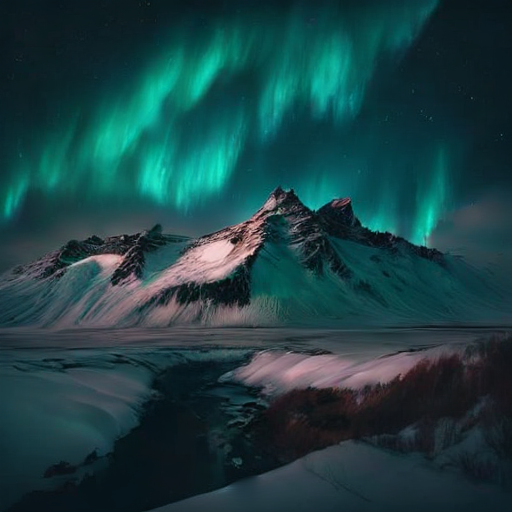}
\\
& \multicolumn{9}{l}{\scriptsize\textit{\quad Happy dreamy owl monster sitting on a tree branch, colorful glittering particles, forest background, detailed feathers.}}\\
& \hspace{2pt}\includegraphics[width=.10\textwidth]{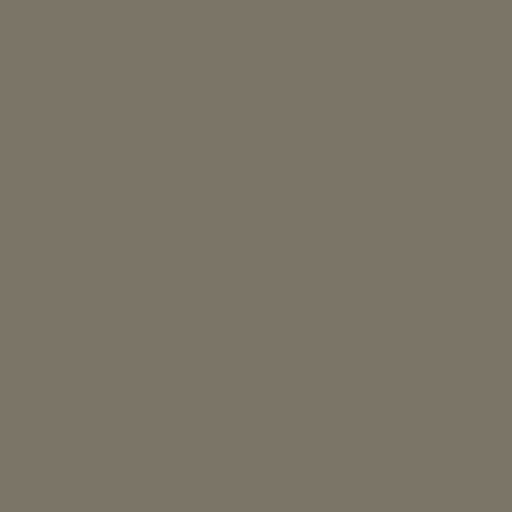}
&\includegraphics[width=.10\textwidth]{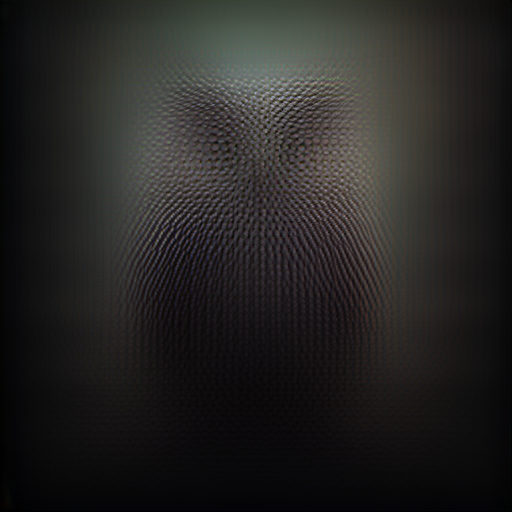}
&\hspace{2pt}\includegraphics[width=.10\textwidth]{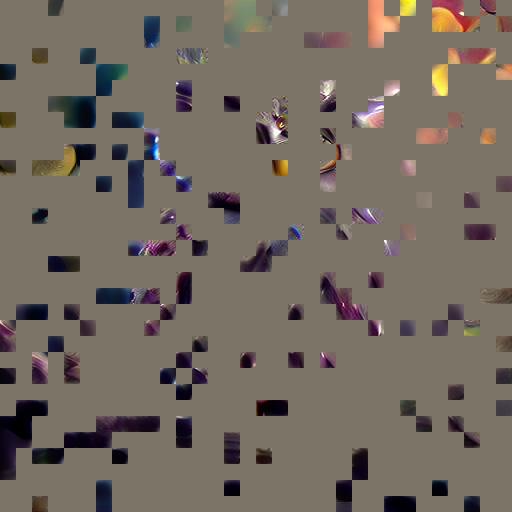}
&\includegraphics[width=.10\textwidth]{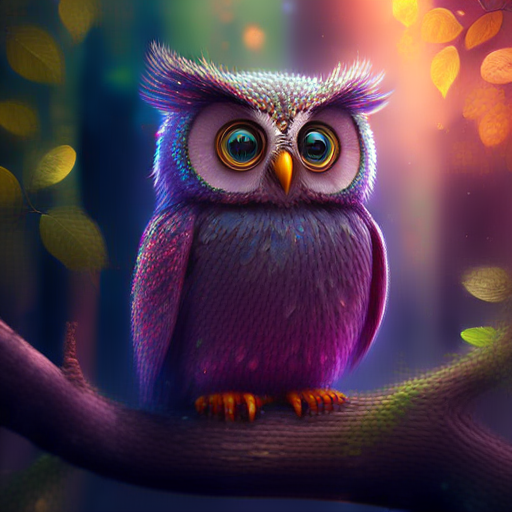}
&\hspace{2pt}\includegraphics[width=.10\textwidth]{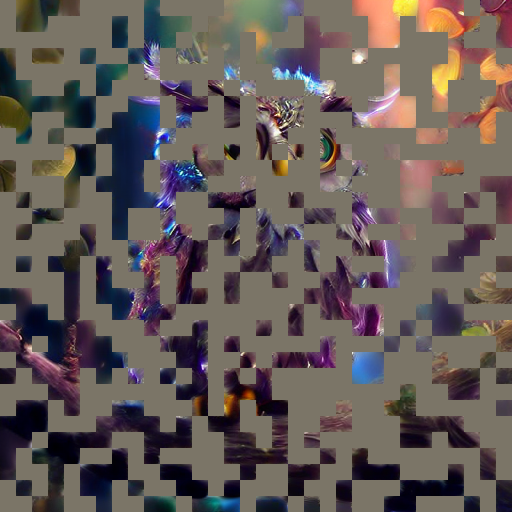}
&\includegraphics[width=.10\textwidth]{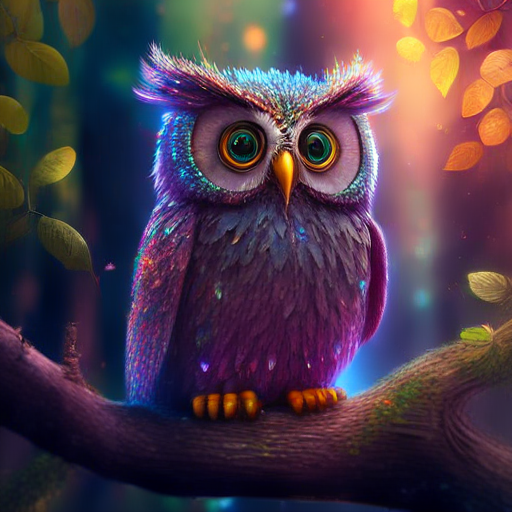}
&\hspace{2pt}\includegraphics[width=.10\textwidth]{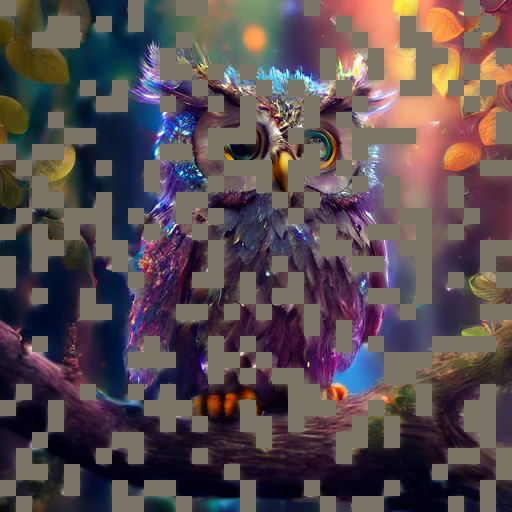}
&\includegraphics[width=.10\textwidth]{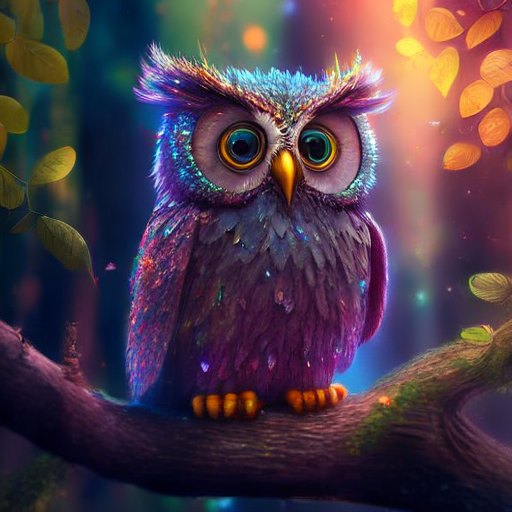}
&\hspace{2pt}\includegraphics[width=.10\textwidth]{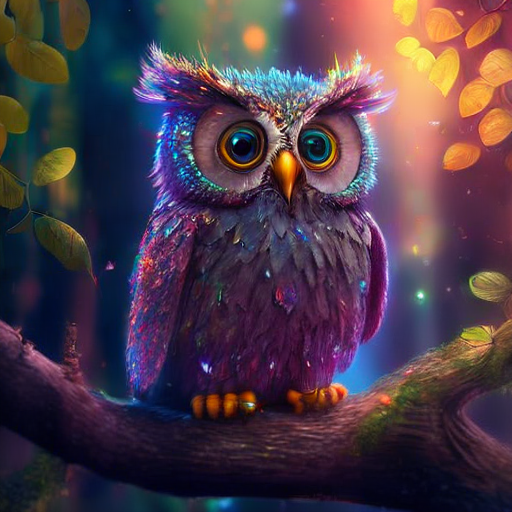}
\\
& \multicolumn{2}{c}{AR Step 0} & \multicolumn{2}{c}{AR Step 26} & \multicolumn{2}{c}{AR Step 42} &\multicolumn{2}{c}{AR Step 53} & Generated\\
\\
\end{tabular}
\caption{\textbf{Image prediction results} at different stages of generation. In each image pair, the left image shows the currently generated tokens, while the right shows the final image we predict based on the generated tokens.
The prediction results are inaccurate and lack details in early stages but become increasingly accurate as more tokens are generated. This is consistent across the four models.}
\label{fig:probing_results_diff_gen_stages}
\end{figure*}

We provide three pieces of empirical evidence to our finding. First, we train a multilayer perceptron (MLP) or repurpose the original model head, based on the hidden representation of generated tokens, to predict the outcomes of the diffusion process. As shown in \figref{fig:probing_results_diff_gen_stages}, in early stages of generation, the MLP prediction is inaccurate and lacks details. In comparison, as more tokens are generated, MLP prediction becomes increasingly accurate, indicating that the autoregressive model now provides stronger conditions for the diffusion head. Second, variance in diffusion sampling gradually decreases during generation, indicating that the distribution of the next token becomes increasingly constrained. Third, based on the straightness metric~\cite{rect_flow}, we show that denoising paths from noise to tokens become closer to straight lines, suggesting that we could take larger step sizes. 

The above finding dictates that fewer diffusion steps are needed in late generation stages than in early stages, forming the proposal of the diffusion step annealing (DiSA) method. Instead of using the same number of diffusion steps throughout the generation process, DiSA uses more diffusion steps (\emph{e.g.}, 50) for early tokens and gradually fewer steps (\emph{e.g.}, from 50 to 5) for later tokens. 

DiSA is training-free and can be easily implemented on top of existing autoregressive diffusion models that share similar token generation mechanisms, such as those in \figref{fig:framework}(a-d). Moreover, because DiSA comes from our finding specific to diffusion in autoregressive models, it can be effectively used together with existing acceleration methods specifically designed for diffusion. Experiments show that DiSA is very useful: it consistently improves the inference efficiency of MAR by $5-10\times$ and FlowAR and xAR by $1.4-2.5\times$ without sacrificing image generation quality. 

In summary, this paper covers three main points. {First}, we reveal that the role of diffusion in autoregressive models is different along the generation process. {Second}, based on this insight, we design a new sampling strategy, DiSA, for scheduling diffusion steps in autoregressive image generation. Third, experiments demonstrate that DiSA delivers very useful acceleration during inference while exhibiting competitive or better generation performance.

\section{Related-Work}\label{sec:related_work}
\textbf{Autoregressive models meet diffusion}. A common practice for autoregressive image generation is to quantize an image into discrete tokens~\cite{vqvae2,vqgan,lee2022autoregressive} and train autoregressive models on the tokens. 
A main bottleneck for these models is that discrete tokens introduce quantization errors, limiting the generation quality~\cite{tschannen2023givt,mar,han2024infinity}. 
To address this, MAR~\cite{mar} uses continuous tokens and adopts a diffusion model head to sample the next tokens in autoregressive models. Other continuous-token design appears later \cite{ren2024flowar,xar,harmon}. 
These methods have good generation quality but low efficiency. 

\textbf{Acceleration techniques for diffusion models}. It is a well-established area in diffusion. Fast sampling processes have been proposed, such as DDIM~\cite{ddim}, DPM-Solver~\cite{lu2022dpm}, and DPM-Solver++~\cite{dpm_plus_plus}, to name a few. These methods are designed specifically for diffusion and can be used together with our approach.
In comparison, less attention has been paid to accelerating diffusion in autoregressive models. LazyMAR \cite{lazymar} introduces two caching techniques, while CSpD~\cite{speculative_mar} applies speculative decoding for speeding up the inference of MAR. These works mainly focus on the autoregressive part of MAR, without modifying the diffusion process, so are orthogonal to our approach. Besides, FAR~\cite{far} replaces the diffusion head of MAR with a short-cut model, achieving 2.3$\times$ acceleration. FAR is trained from scratch, while our method is training-free.

\section{Observation on Autoregressive + Diffusion Models}

\subsection{Revisiting Existing Models}
With an image tokenizer, an image can be represented as a sequence of tokens $\langle \bx^1,\bx^2,\dots,\bx^n\rangle$. For example, we can use VAE~\cite{vae,ldm} to encode an image to 256 tokens. Image generation can be framed as sampling from the joint distribution of image tokens $p(\bx^1,\bx^2,\dots,\bx^n)$. Sampled tokens are decoded by the tokenizer back into images.

An autoregressive model formulates the generation of an image as a next-token prediction task:
\begin{equation}
    p(\bx^1,\bx^2,\dots,\bx^n)=\prod_{i=1}^{n}p(\bx^i \mid \bx^1, \dots, \bx^{i-1}) \text{ where }\bx^i\sim p(\bx^i \mid \bx^1, \dots, \bx^{i-1}).
\end{equation}
Note that recent works propose new autoregressive paradigms~\cite{var,xar}. In next-scale prediction~\cite{var}, given the tokens of a low-resolution image, the model generates the tokens of higher resolution. MAR and xAR generate a group of tokens in each autoregressive step. For these models, $\bx^i$ represents a group of tokens. We interchangeably use $\bx^i$ to denote these tokens for simplicity.

Recent autoregressive models adopt a diffusion process to sample $\bx^i\sim p(\bx^i \mid \bx^1, \dots, \bx^{i-1})$.

\textbf{MAR}~\cite{mar} uses an encoder-decoder backbone $\mathbf{f}$, which takes tokens previously generated as input and predicts a condition vector $\bz^i = \mathbf{f}(\bx^1,\bx^2,\dots,\bx^{i-1})$ for the next token. The diffusion model head, conditional on $\bz^i$, denoises sampled noise to a token via reverse process.

At training time, parameters in $\bepsilon_\btheta$ and $\mathbf{f}$ are updated based on the diffusion loss~\cite{ddpm,improved_ddpm}:
\begin{equation}
\mathcal{L}(\bz^i, \bx^i) = \mathbb{E}_{i,\bepsilon, t} \left[ \left\| \bepsilon - \bepsilon_\btheta(\bx^i_t \mid t, \bz^i) \right\|^2 \right], \bx^i_t = \sqrt{\bar{\alpha}_t} \bx^i + \sqrt{1 - \bar{\alpha}_t} \bepsilon,
\label{eq:diff_loss_eps}
\end{equation}
where $\bepsilon \in \mathbb{R}^d$ is a noise vector sampled from $\mathcal{N}(\mathbf{0}, \bI)$ and $t\sim \mathcal{U}({1,\dots,T})$. $\mathcal{N}$ and $\mathcal{U}$ are Gaussian and Uniform distributions, respectively. $\bar{\alpha}_t$ defines a noise schedule~\cite{ddpm,improved_ddpm}. 

\textbf{FlowAR}~\cite{rect_flow} uses VAR~\cite{var} as the backbone $\mathbf{f}$ and flow matching~\cite{rect_flow,sit} as the the model head $\bv_\btheta$. Similar to MAR, the backbone $\mathbf{f}$ takes previous generated tokens as input, and predicts a condition vector $\bz^i$ for each next token. With a sampled noise token, the flow matching head predicts velocity for denoising the token. At training time, the flow matching loss is calculated as:
\begin{equation}
\mathcal{L}(\bz^i, \bx^i) = \mathbb{E}_{i,\bepsilon\sim \normal{\mathbf{0},\bI}, t\sim[0,1]} \left[ \left\| (\bepsilon - \bx^i) - \bv_\btheta(\bx^i_t \mid t, \bz^i) \right\|^2 \right], \text{where } \bx^i_t = (1-t) \bx^i + t \bepsilon.
\label{eq:flow_matching_loss}
\end{equation}

\textbf{xAR}~\cite{xar} takes both previously generated tokens and sampled noise as input. The model runs 50 times for denoising the noise into tokens and continues to sample the next tokens.

\textbf{Harmon}~\cite{harmon} is a unified model for both text-to-image (T2I) and image-to-text generation. This study focuses on its T2I ability. The backbone in Harmon takes the text prompt and generated tokens as input and produces a condition vector for the next token. A diffusion head, conditional on the vector, denoises sampled noise to the next token.

\subsection{More Tokens Generated, Stronger Constraints on Later Tokens}
The diffusion process in the four models samples the next token from the condition distribution $\bx^i\sim p(\bx^i \mid \bx^1, \dots, \bx^{i-1})$. Our key motivation is that, as more tokens are generated, the condition becomes stronger, making the distribution more constrained and the next tokens easier to sample. We will show empirical evidence to support the motivation.

\underline{First, next tokens can be well predicted at later autoregressive generation stages.} We probe the condition from the generated tokens, \textit{\ie} we use a model to predict the sampled $\bx^i$ based on the hidden representation of the generated tokens $\{\bx^1, \dots, \bx^{i-1}\}$. For MAR and Harmon, we train a MLP model to replace the original model head. The MLP predicts $\bx^i$ directly based on the condition from the generated tokens $\bz^i$. For FlowAR and xAR, we repurpose the original model head for flow matching. Specifically, we feed sampled noise with $t=1$ into the model, obtain the estimated velocity $\bv_\btheta(\bx^i_t \mid t=1, \bz^i)$, and predict the next token as $\bx^i_0=\bx_{t=1}^i-\bv$. Since $\bx_{t=1}^i$ is purely noisy, the model has to directly predict the $\bx^i$ based on $\bz^i$. 

As shown in \figref{fig:probing_results_diff_gen_stages}, in the early stage of generation, the predicted tokens and the generated images are blurry and in low quality. But as more tokens have been generated, the MLP predictions become increasingly more accurate, suggesting that stronger conditions are provided by the generated tokens.

\begin{figure}
\vskip -0.15in
  \centering
  \includegraphics[width=0.90\textwidth,trim=5 440 90 0,clip]{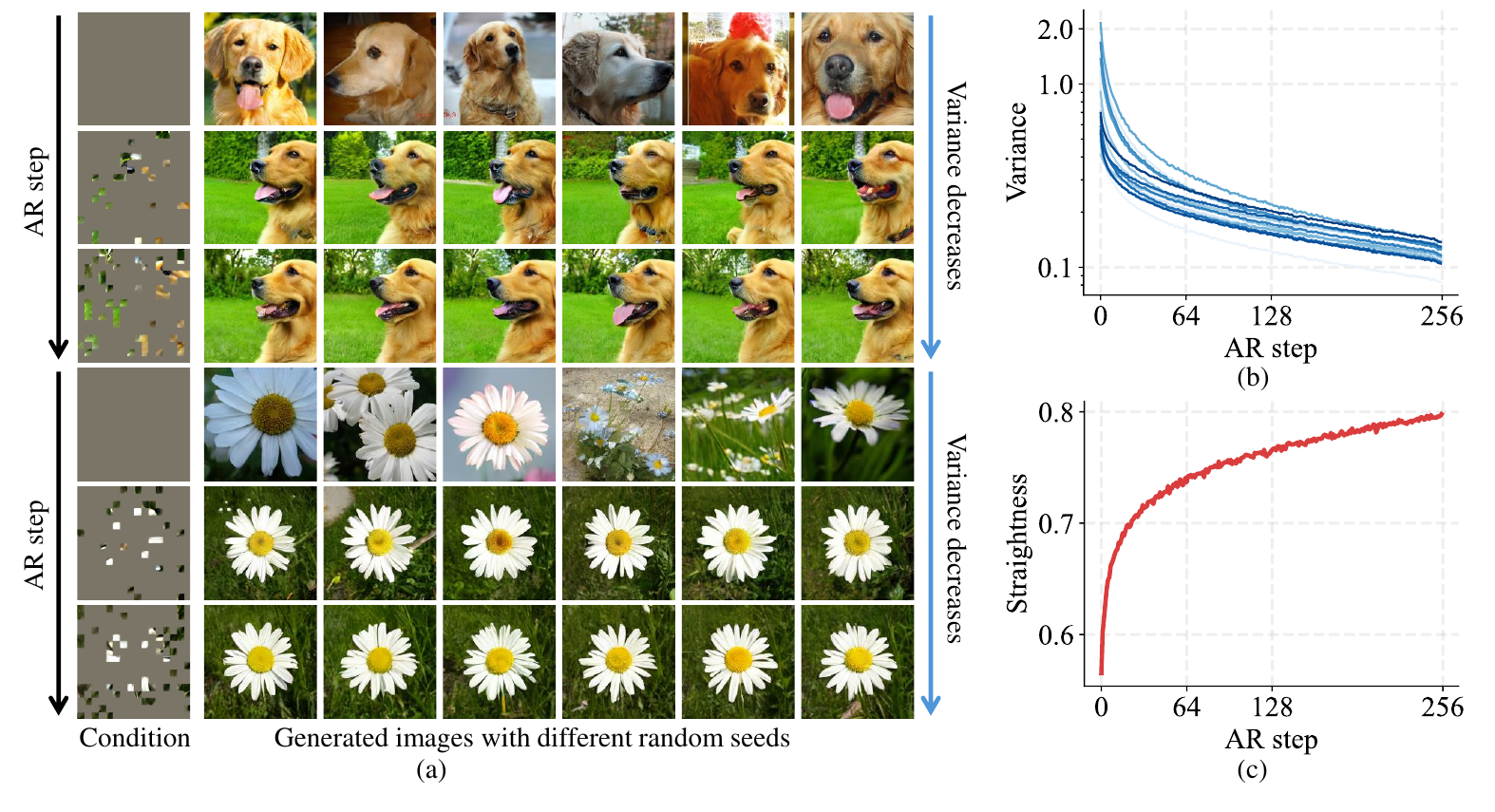}
  \vskip -0.15in
  \caption{\textbf{Diffusion processes in later generation stages} show (a-b) lower variance and (c) closer-to-straight-line denoising paths. (a) Two examples. In each example, the autoregressive step increases from top to bottom rows. 0\%, 10\%, 20\% of tokens have been generated, respectively, as shown in the first column. We observe that the variance of sampled images drops from top to bottom rows. (b) Variance of diffusion-sampled tokens decreases along the autoregressive steps. The y-axis uses a logarithmic scale and each line represents a different token dimension.  (c) Straightness of denoising paths increases from early to late stages. All results are obtained from the MAR-B model.}
  \label{fig:variance}
  \vskip -0.1in
\end{figure}

\underline{Second, next tokens have lower variance at later autoregressive steps.} We explore the variance in the distribution of the next tokens. Specifically, we use MAR to generate 10K images. When generating each $\bx^i$, we sample 100 possible $\bx^i$ and calculate the variance in sampling. The generated examples and the average variance are shown in \figref{fig:variance}(a-b). As seen, as more tokens are generated, the distribution of the next token becomes increasingly constrained. 

\underline{Third, diffusion paths at later stages are closer to straight lines.} Rectified Flow~\cite{rect_flow} proposes that straight paths from noise to data distribution are preferred, because they can be simulated with coarse time discretization and hence need fewer steps at inference time. Inspired by this, we measure the straightness of a denoising path $\{\bx_t\}^{1}_{t=0}$ under condition $\bz$. 
\begin{equation}
    S(\{\bx_t\}^{1}_{t=0}, \bz)=\mathbb{E}_{t\sim [0,1]} \left[ \left\| (\bx_1 - \bx_0)- \bv_\btheta(\bx_t\mid t,\bz) \right\|^2 \right]. \label{eq:straightness_flow}
\end{equation}
MAR and Harmon use diffusion process and are not trained on the rectified flow loss function. Thus, we calculate the cosine similarity between the score (the gradient of the data distribution density)~\cite{song2019generative,song2020score,dhariwal2021diffusion} and the straight direction from the noisy token to the clean token.
\begin{equation}
    S(\{\bx_t\}^{999}_{t=0}, \bz)=\mathbb{E}_{t} \left[ \cos \left(\bx_{0} - \bx_{t}, \nabla_{\bx_t} \log p_\theta (\bx_t\mid t, \bz) \right) \right],
    \label{eq:straightness_diff}
\end{equation}
where $\nabla_{\bx_t} \log p_\theta (\bx_t\mid t, \bz) = -\frac{1}{\sqrt{1 - \bar{\alpha}_t}} \bepsilon_\btheta (\bx_t\mid t, \bz)$. As shown in \figref{fig:variance}(c), in the later stage of generation, the diffusion paths become closer to a stright line, indicating that we can use larger step size and fewer diffusion steps are needed~\cite{rect_flow}. The results on FlowAR, xAR, and Harmon and details of implementation are shown in the Appendix.

\subsection{Diffusion Step Annealing}
Based on the observation, we propose a training-free sampling strategy, DiSA. In the early stage of generation, the distribution of the next tokens is diverse so we allow the diffusion process to run more times, \emph{e.g.}, 50 steps. In the later stage, as the distribution of the next token is more constrained, we assign gradually fewer steps to diffusion, \emph{e.g.}, 5 steps. 

We introduce and compare three different time schedulers in DiSA: two-stage, linear, and cosine. Let $T(k)$ denote the number of diffusion steps when the autoregressive step is $k$. $T_{early}$ and $T_{late}$ are two parameters to control the number of steps. In short, the two-stage method is just cutting the generation into the early and late stages. In the early stage, the diffusion process runs $T_{early}$ while in the late stage, runs $T_{late}$ times. The linear and cosine methods transition smoothly from $T_{early}$ to $T_{late}$ in the generation process. Specifically, they are defined as follows,
\begin{align}
    \text{Two-stage: } & T(k)= \begin{cases}
                        T_{early}, & k < K/2\\
                        T_{late},   & \text{otherwise}
                        \end{cases}, \label{eq:two_stage} \\
    \text{Linear: } & T(k)= T_{early} + (T_{late}-T_{early}) \times k/K, \label{eq:linear}\\
    \text{Cosine: } & T(k)= T_{late} + (T_{early}-T_{late}) \times \frac{1}{2}\left(\cos(\frac{k}{K}\pi)+1\right), \label{eq:cosine}
\end{align}
where $K$ is the total number of the autoregressive steps and $T(k)$ is rounded to the nearest integer. 

\begin{figure}
  \centering
  \includegraphics[width=0.98\textwidth,trim=0 200 0 0,clip]{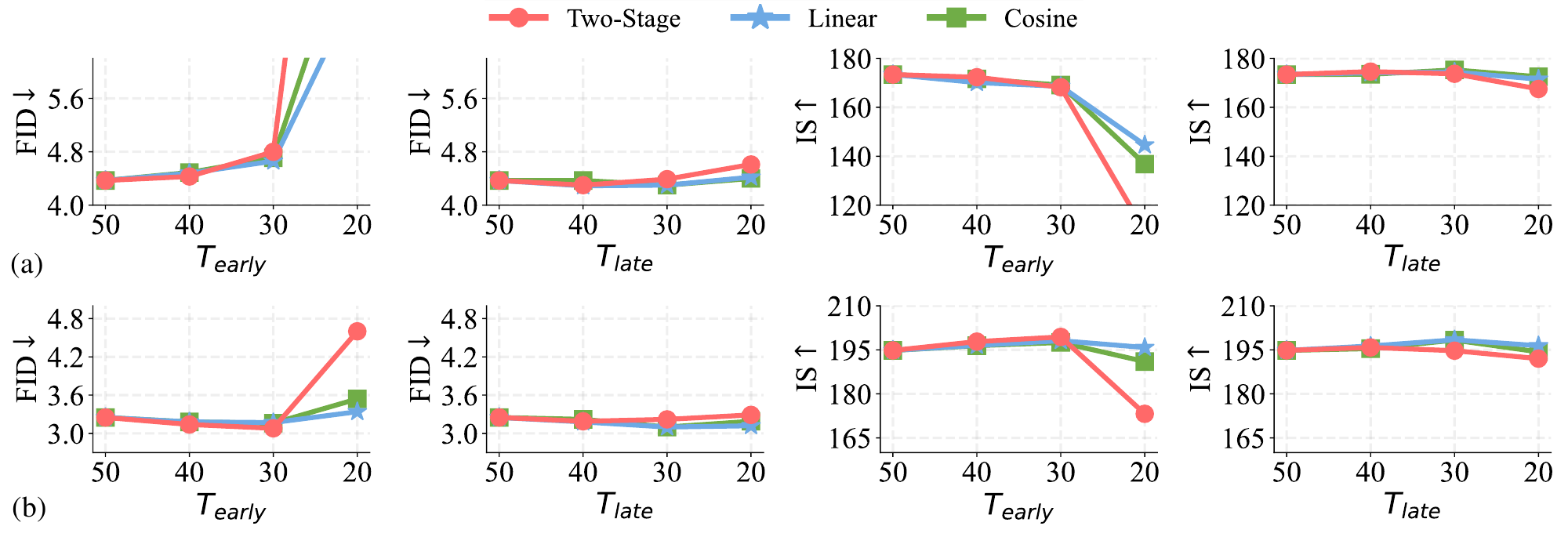}
  \caption{\textbf{Impact of different numbers of diffusion steps} in early generation stages $T_{early}$ and in late stages $T_{late}$ on (a) MAR-B; (b) MAR-L. In the first and third columns, we fix $T_{late}=50$ and reduce $T_{early}$, which significantly degrades generation quality. But as shown in the second and fourth columns, if we fix $T_{early}=50$ and decrease $T_{late}$, the degradation in generation quality is marginal.}
  \label{fig:two-stage}
  \vskip -0.1in
\end{figure}

A preliminary experiment based on MAR is conducted to validate our method. We implement three time schedulers on MAR-B and MAR-L, modify values of $T_{early}$ and $T_{late}$, and evaluate the model on ImageNet 256$\times$256 generation. Fréchet Inception Distance (FID)~\cite{fid} and Inception scores~\cite{is} on 50K sampled images are reported to measure the generation quality. The number of autoregressive step is set to 64, and the default values of $T_{early}$ and $T_{late}$ are both 50. In \figref{fig:two-stage}, reducing number of diffusion steps in early stages degrades the generation quality, but using fewer diffusion steps in later stages does not, which supports our motivation again. We use the linear scheduler in subsequent experiments, which has relatively better performance among the three methods.

We find that reducing $T_{late}$ to less than 20 leads to poor generation results in MAR. The main reason is that the diffusion head has inaccurate prediction around $t=999$. Thus, we let the diffusion start with $t=950$, \ie adding an initial time offset, following the practice in diffusion models~\cite{ddim,pndm,lin2024common,diffusers}. This allows us to further reduce the diffusion steps in MAR. For FlowAR and xAR, we do not observe this phenomenon and the sampling process starts with $t=1.0$. We discuss this further in \secref{sec:main_analysis}.

\subsection{More Insights and Discussions}
\textbf{MAR vs MAE.} We show that an MLP can well predict the remaining tokens. This bridges the underlying mechanism between MAR and masked auto-encoder (MAE)~\cite{he2022masked}. The former uses a generative method to unmask an image, while the latter uses a deterministic way to do so. This is also consistent with recent findings where MAR encodes semantic information for an image~\cite{harmon}.

\textbf{Difficulty level of token distribution modeling.} Condition vectors in later generation stages offer more information, making token distributions easier to model. This may also hold in other autoregressive models. For example, recent works use Gaussian Mixture Model to model token distribution~\cite{tschannen2023givt,zhao2025arinar}. It is possible that the early stage needs more Gaussian components while later stages require fewer. We leave this as future work.

\textbf{Strong diffusion conditions in computer vision.} It is intuitive to understand that the condition vector which summarizes more previously generated tokens is more informative. Therefore, fewer diffusion steps would still sample a good token. This is consistent with some existing works in image contour detection and depth estimation using diffusion models: because of the strong image condition, a few and even one diffusion step would yield competitive results~\cite{liu2025semantic,song2025depthmaster,zhou2024generative}. In T2I generation, a text prompt seems a weak condition. Our prediction results on Harmon in \figref{fig:probing_results_diff_gen_stages} show that, a text prompt helps to determine the basic the structure of the image, leaving details for generation.

\begin{table*}[t]
\vskip -0.1in
\scriptsize 
    \centering
    \caption{\textbf{System-level method comparison} on ImageNet 256×256 
    Our method significantly improves the inference efficiency of MAR, FlowAR, and xAR, while maintaining their generation quality. Diffusion steps ``$a\rightarrow b$'' means starting with $a$ steps and transition to $b$ steps via Eq.~\eqref{eq:linear}. The average inference time per image and speed-ups of different methods are reported.}
    \label{tab:comparison}
    \setlength{\tabcolsep}{1.2mm}{
    \begin{tabular}{l | l | c | c c | c c | c c | l c}
        \toprule
        & Model & \#params & AR steps & Diff steps & {FID$\downarrow$} & {IS$\uparrow$} & {Pre.$\uparrow$} & {Rec.$\uparrow$} & Time (s)$\downarrow$ & Speed-Up$\uparrow$ \\
        \midrule
        \multirow{1.5}[2]{*}{\rotatebox{90}{Diff}}
        & LDM-4$^\dagger$ \cite{Rombach2022} & 400M & - & - & 3.60 & 247.7 & 0.87 & 0.48 & \quad \quad - & - \\
        & DiT-XL/2~\cite{Peebles2023} & 675M & - & 250 & 2.27 & 278.2 & 0.83 & 0.57 & \quad 1.859 & - \\
        
        \midrule
        \multirow{15}[23]{*}{\rotatebox{90}{AR}} & GIVT~\cite{tschannen2023givt} & 304M & 256 & - & 3.35 & - & 0.84 & 0.53 & \quad \quad - & - \\        
        & MAR-B~\cite{mar} & 208M & 256 & 100 & 2.31 & 281.7 & 0.82 & 0.57 & \quad 0.650 & 1.0$\times$ \\
        & & & 64 & 50 & 2.39 (\change{+0.08}) & 281.0 (\change{-0.7}) & 0.82 & 0.57 & \quad 0.134 & 4.8$\times$ \\
        
        & LazyMAR-B~\cite{lazymar} & 208M & 64 & 100 & 2.45 (\change{+0.14}) & 281.3 (\change{-0.4}) & - & - & \quad 0.061$^*$ & 10.6$\times$ \\
        & & & 32 & 100 & 2.64 (\change{+0.33}) & 276.0 (\change{-5.7}) & \quad \quad - & - & \quad 0.045$^*$ & 14.3$\times$ \\
        & FAR-B~\cite{far} & 172M & 256 & 8 & 2.37 (\change{+0.06}) & 265.5 (\change{-16.2}) & - & - & \quad \quad  - & 2.3$\times$ \\
        
         & \cellcolor{lightorange}MAR-B + DiSA & \cellcolor{lightorange}208M & \cellcolor{lightorange}64 & \cellcolor{lightorange}50$\rightarrow$5 & \cellcolor{lightorange} 2.31 (\change{+0.00}) & 282.3\cellcolor{lightorange} (\change{+0.6}) & \cellcolor{lightorange}0.83 & \cellcolor{lightorange}0.56 & \cellcolor{lightorange}\quad 0.114 & \cellcolor{lightorange}5.7$\times$ \\
         
         & \cellcolor{lightorange} & \cellcolor{lightorange} & \cellcolor{lightorange}32 & \cellcolor{lightorange}25$\rightarrow$5 & \cellcolor{lightorange}2.35 (\change{+0.04}) & \cellcolor{lightorange}282.9 (\change{+1.2}) & \cellcolor{lightorange}0.83 & \cellcolor{lightorange}0.56 & \cellcolor{lightorange}\quad 0.057 & \cellcolor{lightorange}11.3$\times$ \\
         \cline{2-11} \\[-1em]  
         
        & MAR-L~\cite{mar} & 479M & 256 & 100 & 1.78 & 296.0 & 0.81 & 0.60 & \quad 1.102 & 1.0$\times$ \\
        & & & 64 & 50 & 1.86 (\change{+0.08}) & 294.0 (\change{-2.0}) & 0.80 & 0.61 & \quad 0.250 & 4.4$\times$ \\

        & LazyMAR-L~\cite{lazymar} & 479M & 64 & 100 & 1.93 (\change{+0.15}) & 297.4 (\change{+1.4}) & - & - & \quad 0.106$^*$ & 10.4$\times$ \\
        & & & 32 & 100 & 2.11 (\change{+0.33}) & 284.4 (\change{-11.6}) & - & - & \quad 0.080$^*$ & 13.8$\times$ \\

        & FAR-L~\cite{far} & 406M & 256 & 8 & 1.99 (\change{+0.21}) & 293.0 (\change{-3.0}) & - & - & \quad \quad - & 1.4$\times$ \\

        & MAR-L + CSpD~\cite{speculative_mar} & - & - & - & 1.81 (\change{+0.03}) & 303.7 (\change{+7.7}) &- &- & \quad \quad - & 1.5$\times$ \\
        
        & \cellcolor{lightorange}MAR-L + DiSA & \cellcolor{lightorange}479M & \cellcolor{lightorange}64 & \cellcolor{lightorange}50$\rightarrow$5 & \cellcolor{lightorange}1.77 (\change{-0.01}) & \cellcolor{lightorange}298.3 (\change{+2.3}) & \cellcolor{lightorange}0.81 & \cellcolor{lightorange}0.61 & \cellcolor{lightorange}\quad 0.216 & \cellcolor{lightorange}5.1$\times$ \\
        
        & \cellcolor{lightorange} & \cellcolor{lightorange} & \cellcolor{lightorange}32 & \cellcolor{lightorange}25$\rightarrow$5 & \cellcolor{lightorange} 1.88 (\change{+0.10}) & \cellcolor{lightorange}295.1 (\change{-0.9}) & \cellcolor{lightorange}0.81 & \cellcolor{lightorange}0.61 & \cellcolor{lightorange}\quad 0.108 & \cellcolor{lightorange}10.2$\times$ \\
        \cline{2-11}\\[-1em]  
        
        & MAR-H~\cite{mar} & 943M & 256 & 100 & 1.55 & 303.7 & 0.81 & 0.62 & \quad 1.957 & 1.0$\times$ \\
        & & & 64 & 50 & 1.65 (\change{+0.10}) & 299.8 (\change{-3.9}) & 0.80 & 0.62 & \quad 0.462 & 4.2$\times$ \\

        & LazyMAR-H~\cite{lazymar} & 943M & 64 & 100 & 1.69 (\change{+0.14}) & 299.2 (\change{-4.5}) & - & - & \quad 0.191$^*$ & 10.2$\times$ \\
        & & & 32 & 100 & 1.94 (\change{+0.39}) & 284.1 (\change{-19.6}) & - & - & \quad 0.145$^*$ & 13.5$\times$ \\
        & MAR-H + CSpD~\cite{speculative_mar} & - & - & - & 1.60 (\change{+0.05}) & 301.6 (\change{-2.1}) &- &- & \quad \quad - & 2.3$\times$ \\
        
        & \cellcolor{lightorange}MAR-H + DiSA & \cellcolor{lightorange}943M & \cellcolor{lightorange}64 & \cellcolor{lightorange}50$\rightarrow$5 & \cellcolor{lightorange}1.57 (\change{+0.02}) & \cellcolor{lightorange}303.1 (\change{-0.6}) & \cellcolor{lightorange}0.80 & \cellcolor{lightorange}0.62 & \cellcolor{lightorange}\quad 0.404 &\cellcolor{lightorange}4.8$\times$ \\
        
        & \cellcolor{lightorange} & \cellcolor{lightorange} & \cellcolor{lightorange}32 & \cellcolor{lightorange}25$\rightarrow$5 & \cellcolor{lightorange}1.72 (\change{+0.17}) & \cellcolor{lightorange}303.4 (\change{-0.3}) & \cellcolor{lightorange}0.80 & \cellcolor{lightorange}0.61 & \cellcolor{lightorange}\quad 0.209 & \cellcolor{lightorange}9.3$\times$ \\

        \midrule
        \multirow{4}[7]{*}{\rotatebox{90}{VAR}}
        & VAR-d30~\cite{tian2025var} & 2.0B & 10 & - & 1.92 & 323.1 & 0.82 & 0.59 & \quad 0.039$^\dagger$ & -\\
        & FlowAR-S~\cite{ren2024flowar} & 170M & 5 & 25 & 3.70 & 235.1 & 0.81 & 0.51 & \quad 0.024 & 1.0$\times$ \\
        & \cellcolor{lightorange}FlowAR-S + DiSA & \cellcolor{lightorange} 170M & \cellcolor{lightorange}5 & \cellcolor{lightorange}25$\rightarrow$15 & \cellcolor{lightorange}3.74 (\change{+0.04}) & \cellcolor{lightorange}235.2 (\change{+0.01}) & \cellcolor{lightorange}0.81 & \cellcolor{lightorange}0.51 & \cellcolor{lightorange}\quad 0.018 & \cellcolor{lightorange}1.4$\times$ \\
        
        & FlowAR-L~\cite{ren2024flowar} & 589M & 5 & 25 & 1.87 & 273.1 & 0.80 & 0.62 & \quad 0.124 & 1.0$\times$ \\
        & \cellcolor{lightorange}FlowAR-L + DiSA & \cellcolor{lightorange}589M & \cellcolor{lightorange}5 & \cellcolor{lightorange}25$\rightarrow$15 & \cellcolor{lightorange}1.90 (\change{+0.03}) & \cellcolor{lightorange}274.8 (\change{+1.7}) & \cellcolor{lightorange}0.80 & \cellcolor{lightorange}0.61 & \cellcolor{lightorange}\quad 0.082 & \cellcolor{lightorange}1.5$\times$ \\

        & FlowAR-H~\cite{ren2024flowar} & 1.9B & 5 & 50 & 1.67 & 276.3 & 0.80 & 0.62 & \quad 0.423$^\dagger$ & 1.0$\times$ \\
        & \cellcolor{lightorange}FlowAR-H + DiSA & \cellcolor{lightorange}1.9B & \cellcolor{lightorange}5 & \cellcolor{lightorange}50$\rightarrow$15 & \cellcolor{lightorange}1.69 (\change{+0.02}) & \cellcolor{lightorange}273.8 (\change{-2.5}) & \cellcolor{lightorange}0.80 & \cellcolor{lightorange}0.62 & \cellcolor{lightorange}\quad 0.167$^\dagger$ & \cellcolor{lightorange}2.5$\times$ \\
        
        \midrule
        \multirow{5}[4]{*}{\rotatebox{90}{xAR}} & xAR-B~\cite{xar} & 172M & 4 & 50 & 1.67 & 265.2 & 0.80 & 0.62 & \quad 0.130 & 1.0$\times$ \\
        
        & \cellcolor{lightorange}xAR-B + DiSA & \cellcolor{lightorange}172M & \cellcolor{lightorange}4 & \cellcolor{lightorange}50$\rightarrow$15 & \cellcolor{lightorange}1.68 (\change{+0.01}) & \cellcolor{lightorange}265.5 (\change{+0.3}) & \cellcolor{lightorange}0.79 & \cellcolor{lightorange}0.62 & \cellcolor{lightorange}\quad 0.084 &\cellcolor{lightorange}1.6$\times$ \\
        
        & xAR-L~\cite{xar} & 608M & 4 & 50 & 1.28 & 292.5 & 0.82 & 0.62 & \quad 0.394 & 1.0$\times$ \\
        & \cellcolor{lightorange}xAR-L + DiSA & \cellcolor{lightorange}608M & \cellcolor{lightorange}4 & \cellcolor{lightorange}50$\rightarrow$15 & \cellcolor{lightorange}1.23 (\change{-0.05}) & \cellcolor{lightorange}287.3 (\change{-5.2}) & \cellcolor{lightorange}0.79 & \cellcolor{lightorange}0.66 & \cellcolor{lightorange}\quad 0.255 &\cellcolor{lightorange}1.5$\times$ \\
        
        & xAR-H~\cite{xar} & 1.1B & 4 & 50 & 1.24 & 301.6 & 0.83 & 0.64 & \quad 0.896$^\dagger$ & 1.0$\times$ \\
        & \cellcolor{lightorange}xAR-H + DiSA & \cellcolor{lightorange}1.1B & \cellcolor{lightorange}4 & \cellcolor{lightorange}50$\rightarrow$15 & \cellcolor{lightorange}1.23 (\change{-0.01}) & \cellcolor{lightorange}300.5 (\change{-1.1}) & \cellcolor{lightorange}0.79 & \cellcolor{lightorange}0.66 & \cellcolor{lightorange}\quad 0.577$^\dagger$ &\cellcolor{lightorange}1.6$\times$ \\
        \bottomrule
    \end{tabular}}

\raggedright$^\dagger$ We test the latency of generating a batch of 128 images instead of 256 to reduce memory usage. \raggedright$^*$ Estimated based on their paper.
\vskip -0.2in
\end{table*}

\section{Experiments}
\subsection{Implementation Details, Datasets, and Metrics}
Experiments mainly includes four pretrained models: MAR~\cite{mar}, FlowAR~\cite{ren2024flowar}, xAR~\cite{xar}, and Harmon~\cite{harmon}. MAR, FlowAR, and xAR are evaluated on the ImageNet $256\times256$ generation task. We report FID~\cite{fid}, IS~\cite{is}, Precision, and Recall, following common practice in image generation~\cite{dhariwal2021diffusion}. We also measure the inference time of generating a batch of 256 images for these models. Harmon is evaluated on the T2I benchmark GenEval~\cite{ghosh2023geneval}. Averaged accuracy and inference time are reported. All experiments are run on 4 NVIDIA A100 PCIe GPUs.

\subsection{Evaluation}\label{sec:main_analysis}
\begin{table}[t]
\scriptsize 
    \centering
    \caption{\textbf{Text-to-image generation} of Harmon on GenEval benchmark. The accuracy on each task and the average inference time per image are reported.}
    \label{tab:t2i_res}
    \setlength{\tabcolsep}{1.5mm}{
    \begin{tabular}{llccccccccc}
    \toprule
AR steps & Diff steps & Single Obj. & Two Obj. & Counting & \textbf{Colors} & Position & Color Attri. & Overall & Time per image (s) \\
\midrule
32 & 50 & 0.99 & 0.86 & 0.64 & 0.87 & 0.43 & 0.49 & 0.71 & 12 \\
& \cellcolor{lightorange}50$\rightarrow$5 & \cellcolor{lightorange}0.99 & \cellcolor{lightorange}0.85 & \cellcolor{lightorange}0.69 & \cellcolor{lightorange}0.86 & \cellcolor{lightorange}0.48 & \cellcolor{lightorange}0.52 & \cellcolor{lightorange}0.73 & \cellcolor{lightorange}8 \\
\midrule
64 & 25 & 0.05 & 0.00 & 0.00 & 0.00 & 0.00 & 0.00 & 0.01 & 17 \\
& \cellcolor{lightorange}25$\rightarrow$5 & \cellcolor{lightorange}0.99 & \cellcolor{lightorange}0.89 & \cellcolor{lightorange}0.74 & \cellcolor{lightorange}0.86 & \cellcolor{lightorange}0.46 & \cellcolor{lightorange}0.54 & \cellcolor{lightorange}0.75 & \cellcolor{lightorange}14\\
& 50 & 0.99 & 0.88 & 0.71 & 0.88 & 0.48 & 0.53 & 0.74 & 24 \\
 & \cellcolor{lightorange}50$\rightarrow$5 & \cellcolor{lightorange}0.99  & \cellcolor{lightorange}0.89  & \cellcolor{lightorange}0.68  & \cellcolor{lightorange}0.87  & \cellcolor{lightorange}0.41  & \cellcolor{lightorange}0.55  & \cellcolor{lightorange}0.73  & \cellcolor{lightorange}17 \\
 & 100 & 0.99 & 0.86 & 0.69 & 0.89 & 0.48 & 0.50 & 0.73 & 40 \\
 & \cellcolor{lightorange}100$\rightarrow$5 & \cellcolor{lightorange}0.99  & \cellcolor{lightorange}0.90  & \cellcolor{lightorange}0.68  & \cellcolor{lightorange}0.86  & \cellcolor{lightorange}0.49  & \cellcolor{lightorange}0.51  & \cellcolor{lightorange}0.74  & \cellcolor{lightorange}24 \\
\bottomrule
    \end{tabular}}
    \vskip -0.15in
\end{table}

\textbf{DiSA consistently improves the effiency of baseline AR+Diffusion models.} 
We apply DiSA to MAR, xAR, and FlowAR and compare the performance on the ImageNet $256\times256$ generation task in \tabref{tab:comparison}. Overall, DiSA  consistently enhances the efficiency of the baseline models while maintaining competitive generation quality. 

For MAR, the original best performance is achieved with 256 autoregressive steps and 100 diffusion steps. After integrating DiSA to MAR, \emph{e.g.,} $50\rightarrow5$, we report speed-up of 5.7$\times$ on MAR-B, 5.1$\times$ on MAR-L, and 4.8$\times$ on MAR-H, respectively. The generation quality change is minor: 
DiSA results in the same FID on MAR-B and increases FID by 0.02 on MAR-H.

If we further reduce MAR to 32 autoregressive steps and $25\rightarrow 5$ diffusion steps, DiSA results in 9.3-11.3$\times$ speed-ups on MAR with slightly degraded generation quality. For example, DiSA achieves 11.3$\times$ faster inference on MAR-B while increasing FID by 0.04. 

Similarly, FlowAR-H with DiSA achieves a 2.5$\times$ speed-up while maintaining a competitive FID of 1.69 and IS of 273.8. In the case of xAR models, DiSA provides up to 1.6$\times$ speed-up with negligible impact on performance metrics. Interestingly, xAR-L shows 1.6$\times$ speed-up and even improved FID from 1.28 to 1.23 with DiSA. These results clearly indicate the usefulness of DiSA.

\begin{wraptable}{t}{0.4\textwidth}
\vspace{-5pt}
\scriptsize 
    \centering
    \caption{Existing methods speed up MAR sampling and can be used together with DiSA for further speed-up. The number of autoregressive steps is 64. }
    \label{tab:other_techniques}
    \setlength{\tabcolsep}{1.2mm}{
    \begin{tabular}{llccc}
    \toprule
Method & \#Steps & FID$\downarrow$ & IS$\uparrow$ & Time (s)$\downarrow$ \\
\midrule
Original & 25 & 6.78 & 148.8 & 17.0 \\
& 50 & 4.30 & 174.5 & 21.9 \\
 & 100 & 4.38 & 173.7 & 30.6 \\
 \midrule
Time offset & 25 & 4.61 & 171.0 & 16.8 \\
 & 50 & 4.64 & 171.1 & 20.7 \\
\cellcolor{lightorange}+ DiSA & \cellcolor{lightorange}50$\rightarrow$5 & \cellcolor{lightorange}4.17 & \cellcolor{lightorange}173.7 & \cellcolor{lightorange}17.0 \\
 \midrule
DDIM & 25 & 4.16 & 178.2 & 17.7 \\
 & 50 & 4.06 & 176.6 & 22.1 \\
 \cellcolor{lightorange}+ DiSA & \cellcolor{lightorange}50$\rightarrow$5 & \cellcolor{lightorange}4.00 & \cellcolor{lightorange}179.3 & \cellcolor{lightorange}17.9 \\
 \midrule
 DPM-Solver & 15 & 4.58 & 179.4 & 17.4 \\
& 25 & 4.35 & 176.1 & 20.6 \\
\cellcolor{lightorange}+ DiSA & \cellcolor{lightorange}25$\rightarrow$10 & \cellcolor{lightorange}4.37 & \cellcolor{lightorange}177.1 & \cellcolor{lightorange}17.9 \\ 
  \midrule
DPM-Solver++ & 15 & 4.57 & 179.5 & 18.5 \\
 & 25 & 4.34 & 176.1 & 22.0 \\
 \cellcolor{lightorange}+ DiSA & \cellcolor{lightorange}25$\rightarrow$10 & \cellcolor{lightorange}4.37 & \cellcolor{lightorange}177.2 & \cellcolor{lightorange}19.0 \\
\bottomrule
    \end{tabular}}
\vspace{-10pt}
\end{wraptable}

\textbf{Comparison with other acceleration methods on MAR.} DiSA is faster than CSpD~\cite{speculative_mar} and FAR~\cite{far}, and is competitive to LazyMAR~\cite{lazymar}. Note that LazyMAR works on caching techniques for MAR, without modifying the diffusion process, and is orthogonal to DiSA. It is interesting to combine LazyMAR and DiSA in future work.

\textbf{DiSA is also useful on T2I generation models.} As shown in \tabref{tab:t2i_res}, DiSA can also speed up Harmon on T2I generation tasks on GenEval. As seen, Harmon with DiSA uses 8 seconds per image, $5\times$ faster than the original implementation, while achieving a comparable performance.

\textbf{DiSA is complementary to existing diffusion acceleration methods.}
We implement several existing diffusion acceleration techniques on MAR-B. \textbf{Time offset:} We start the diffusion process from $t=950$ instead of $t=999$. \textbf{Faster samplers:} We include DDIM~\cite{ddim}, DPM-Solver~\cite{lu2022dpm}, and DPM-Solver++~\cite{dpm_plus_plus}. Note that FlowAR uses the Euler sampler while xAR uses the Euler-Maruyama sampler~\cite{maruyama1955continuous, higham2001algorithmic}, so we omit detailed discussions of the two samplers here.

As shown in \tabref{tab:other_techniques}, existing techniques designed for diffusion can accelerate sampling in autoregressive models. Time offset reduces the number of diffusion steps but suffers from a slight quality degradation. DDIM achieves a remarkable FID of 4.06 at 50 steps and 4.16 at 25 steps. DPM-Solver and DPM-Solver++ show comparable performance and reduce the number of diffusion steps to 25. 

Our method is complementary to these diffusion acceleration approaches. If we combine time offset with DiSA, inference time can be reduced to 17.0 and FID is improved to 4.17. With a similar inference speed, time offset uses 25 steps and FID is 4.61. For the other three solvers, combining with DiSA also improves the inference speed while maintaining a comparable generation quality.

\begin{figure}
  \centering
  \includegraphics[width=0.99\textwidth,trim=12 220 12 0,clip]{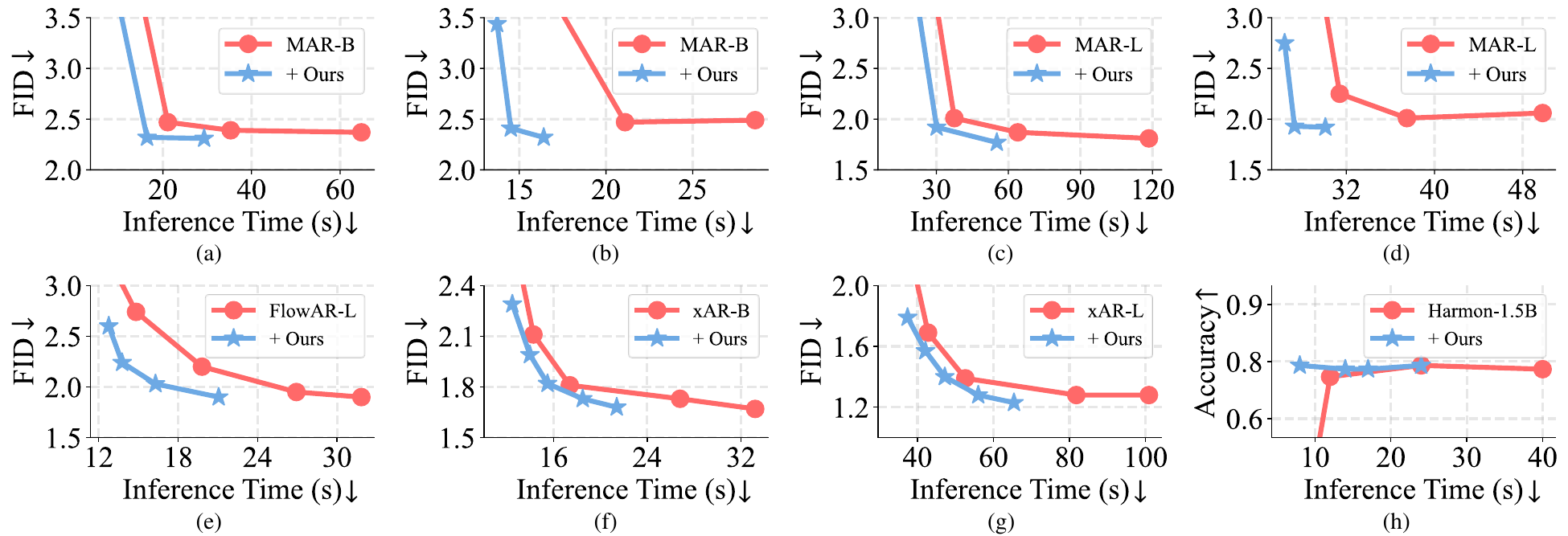}
  \caption{\textbf{Speed-quality trade-off} for (a) MAR-B with \{16, 32, 64, 128\} autoregressive steps; (b) MAR-B with \{25, 50, 100\} diffusion steps; (c) MAR-L with \{16, 32, 64, 128\} autoregressive steps; (d) MAR-L with \{25, 50, 100\} diffusion steps; (e) FlowAR-L with \{8, 10, 15, 20, 25 \} flow matching steps; (f) xAR-B and (g) xAR-L with \{15, 20, 25, 40, 50\} flow matching steps; and (h) Harmon-1.5B with different autoregressive and diffusion steps. }
  \label{fig:trade-off}
\vskip -0.1in
\end{figure}

\begin{figure*}
\centering
\renewcommand\arraystretch{0.6}
\renewcommand\tabcolsep{0pt}
\scriptsize
\begin{tabular}{lcccccccc}
\rotatebox{90}{\quad\ \ \ \  MAR-H} &
\hspace{2pt}\includegraphics[width=.12\textwidth]{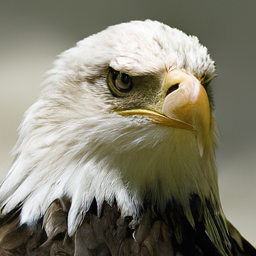}
&\includegraphics[width=.12\textwidth]{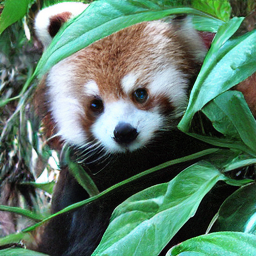}
&\hspace{2pt}\includegraphics[width=.12\textwidth]{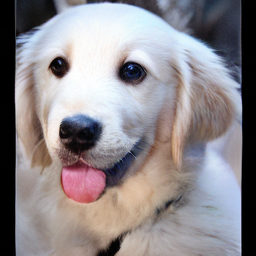}
&\includegraphics[width=.12\textwidth]{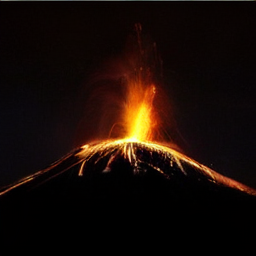}
&\hspace{2pt}\includegraphics[width=.12\textwidth]{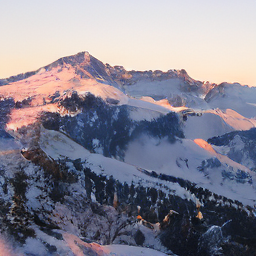}
&\includegraphics[width=.12\textwidth]{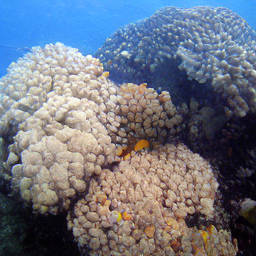}
&\hspace{2pt}\includegraphics[width=.12\textwidth]{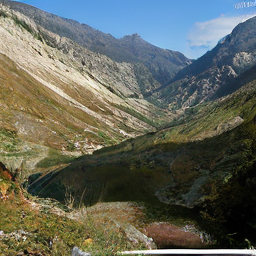}
&\includegraphics[width=.12\textwidth]{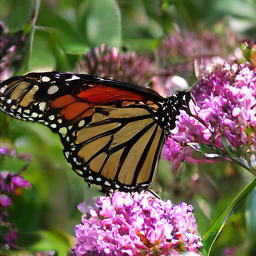}
\\
\rotatebox{90}{\ \ \ Harmon-1.5B} &
\hspace{2pt}\includegraphics[width=.12\textwidth]{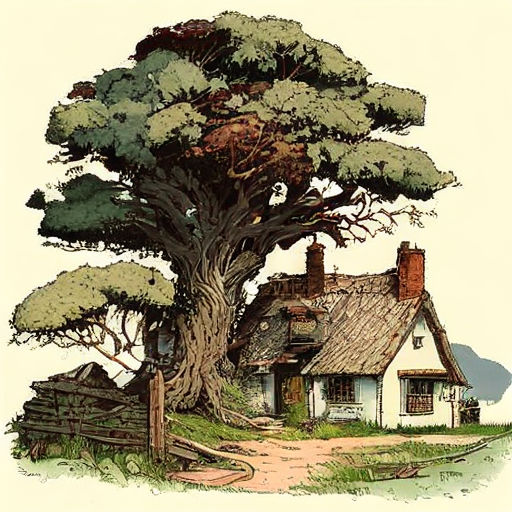}
&\includegraphics[width=.12\textwidth]{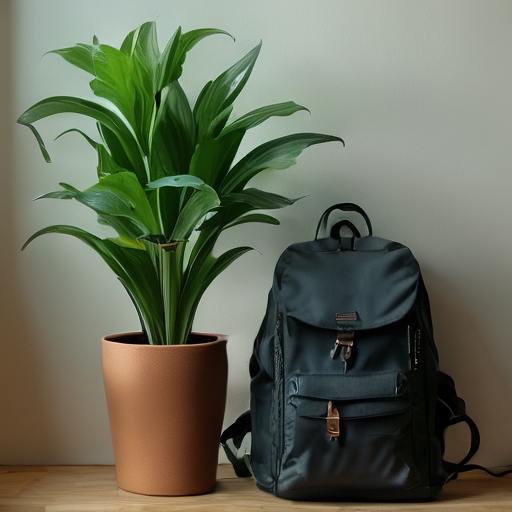}
&\hspace{2pt}\includegraphics[width=.12\textwidth]{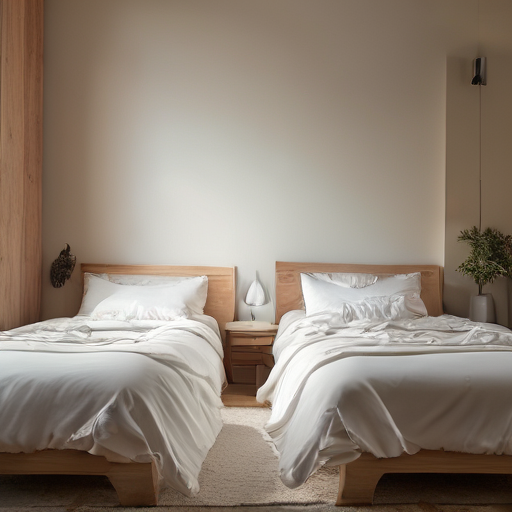}
&\includegraphics[width=.12\textwidth]{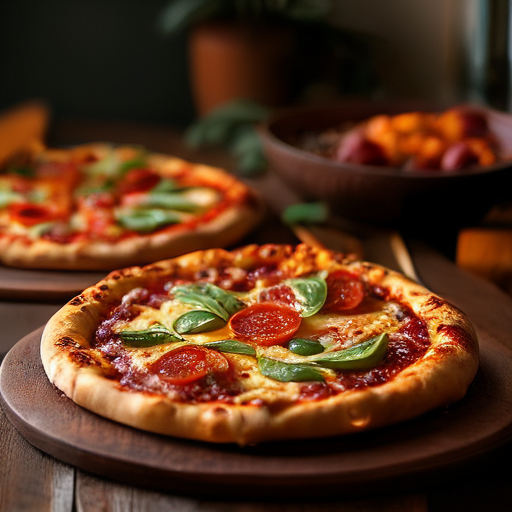}
&\hspace{2pt}\includegraphics[width=.12\textwidth]{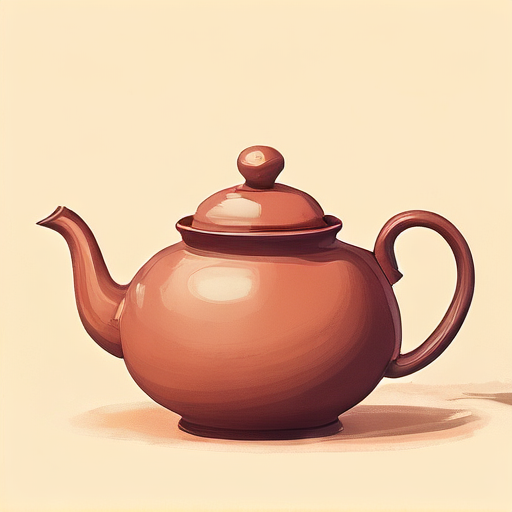}
&\includegraphics[width=.12\textwidth]{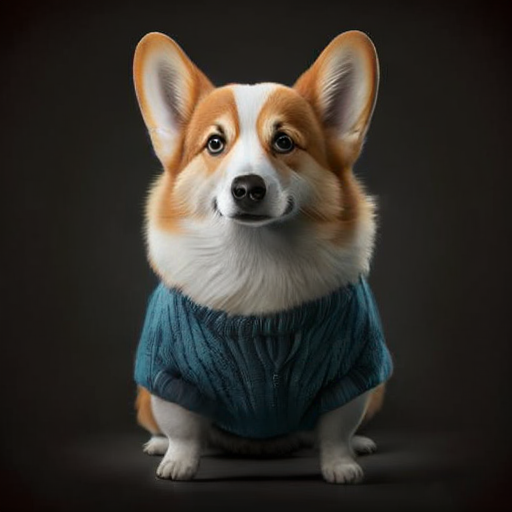}
&\hspace{2pt}\includegraphics[width=.12\textwidth]{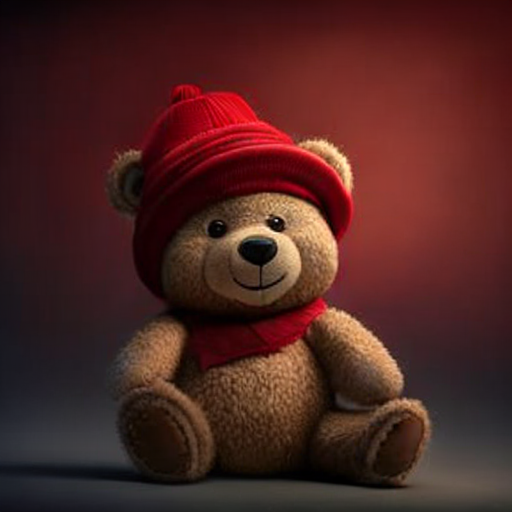}
&\includegraphics[width=.12\textwidth]{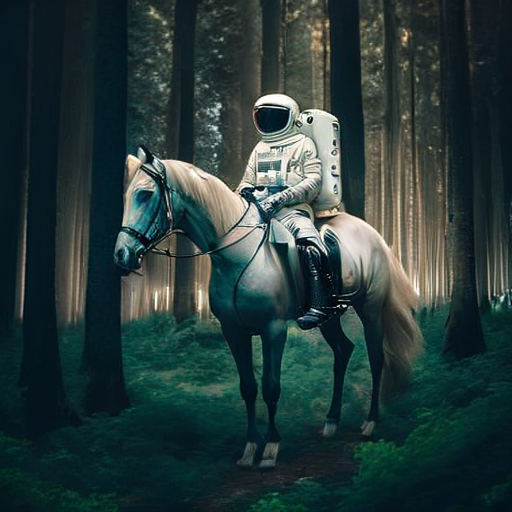}
\\
\rotatebox{90}{\quad FlowAR-H} &
\hspace{2pt}\includegraphics[width=.12\textwidth]{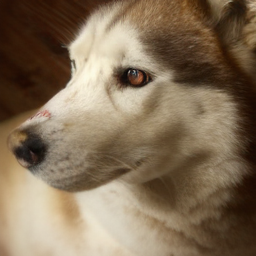}
&\includegraphics[width=.12\textwidth]{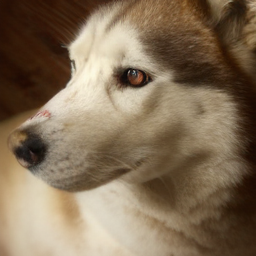}
&\hspace{2pt}\includegraphics[width=.12\textwidth]{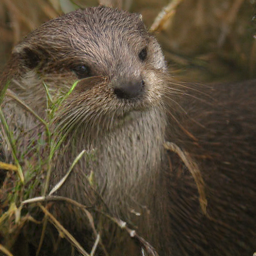}
&\includegraphics[width=.12\textwidth]{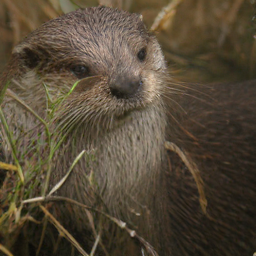}
&\hspace{2pt}\includegraphics[width=.12\textwidth]{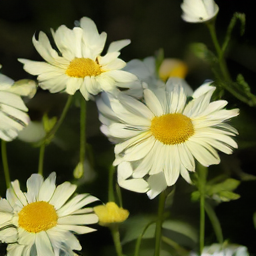}
&\includegraphics[width=.12\textwidth]{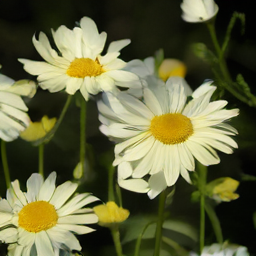}
&\hspace{2pt}\includegraphics[width=.12\textwidth]{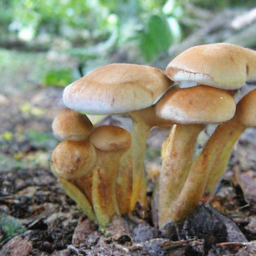}
&\includegraphics[width=.12\textwidth]{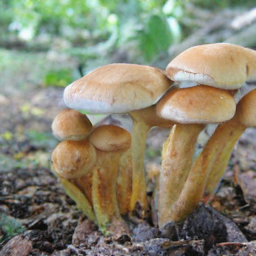}
\\
\rotatebox{90}{\quad\quad xAR-H} &
\hspace{2pt}\includegraphics[width=.12\textwidth]{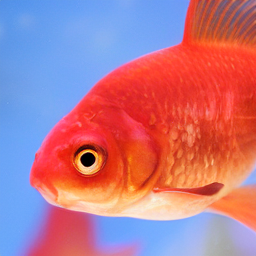}
&\includegraphics[width=.12\textwidth]{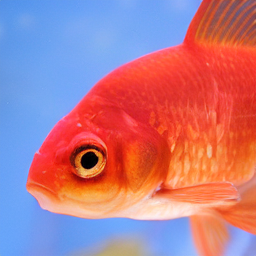}
&\hspace{2pt}\includegraphics[width=.12\textwidth]{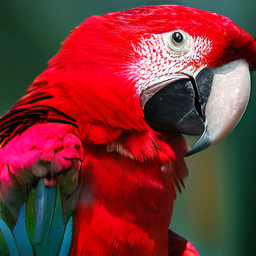}
&\includegraphics[width=.12\textwidth]{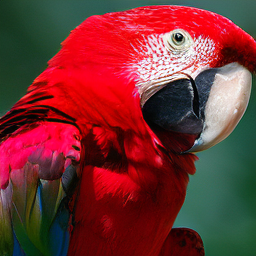}
&\hspace{2pt}\includegraphics[width=.12\textwidth]{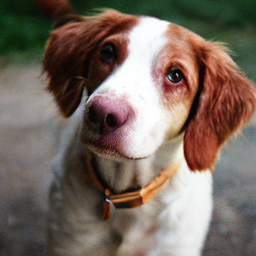}
&\includegraphics[width=.12\textwidth]{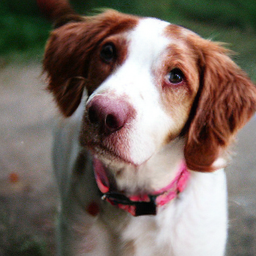}
&\hspace{2pt}\includegraphics[width=.12\textwidth]{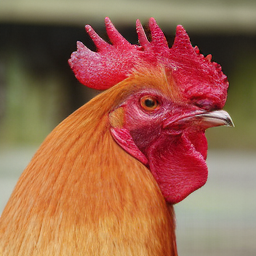}
&\includegraphics[width=.12\textwidth]{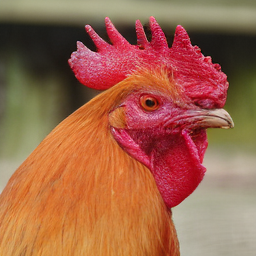}
\\
\end{tabular}
\caption{\textbf{Sample image generation results.} For MAR-H and Harmon-1.5B, we present the samples generated using DiSA. For FlowAR and xAR, each image pair is generated with the same random seed, where the first is generated without DiSA while the other is with DiSA. 
We find that DiSA helps generate similar quality images while speeding up image generation by $2.5\times$ and $1.6\times$ respectively.}
  \label{fig:examples}
\vskip -0.15in
\end{figure*}

\textbf{Trade-off between efficiency and quality.} 
We show the trade-off of speed and generation quality in \figref{fig:trade-off}. For MAR-B and MAR-L, we evaluate different autoregressive and diffusion steps. FlowAR-L, xAR-B, and xAR-L are evaluated with different flow matching steps. Harmon-1.5B runs with different autoregressive and diffusion steps on the GenEval benchmark. As seen, under different settings, DiSA can significantly improve the inference speed of these models, while maintaining the generation quality. We also present sample generation results in \figref{fig:examples}. Detailed results and more examples are provided in the Appendix. 

\section{Conclusion}
W study how to effectively reduce the number of diffusion steps in autoregressive models. We find that as more tokens are generated, the reliance on many diffusion steps is alleviated. Based on this, we propose DiSA, a training-free strategy that gradually decreases the number of diffusion steps during the generation process. This approach is easy to implement and significantly improves inference speed while maintaining competitive image quality. Our study provides interesting insights into the diffusion process in autoregressive image generation, and our future work will focus on how perception models and generative models converge.

\section*{Acknowledgments}
We would like to express our sincere appreciation to Tianhong Li, Zhanhao Liang, and Zhengyang Yu for the insightful discussion that greatly inspired our thinking during the course of this project. We are also deeply thankful to Caixia Zhou, Xingjian Leng, Sam Bahrami, Francis Snelgar, Qingtao Yu, Yunzhong Hou, Weijian Deng, Yang Yang, Yuchi Liu, Zeyu Zhang, and all our lab colleagues for their invaluable support. Their collaborative efforts, insightful discussion, and constructive feedback have been crucial in shaping and improving our paper. This work was supported by an Australian Research Council (ARC) Linkage grant (project number LP210200931).

{\small
\bibliographystyle{plain}
\bibliography{reference}
}



\end{document}